%% file: IEEE-conference-template-062824.tex
\PassOptionsToPackage{table}{xcolor} 
\documentclass[conference]{IEEEtran}
\IEEEoverridecommandlockouts

\usepackage{cite}
\usepackage{amsmath,amssymb,amsfonts}
\usepackage{graphicx}
\usepackage{tabularx}
\usepackage{graphicx} 
\usepackage{multirow} 
\usepackage{enumitem} 
\usepackage{graphicx}
\usepackage{graphics}
\usepackage{graphicx} 
\usepackage{tablefootnote}
\usepackage{amssymb}  
\usepackage[normalem]{ulem}
\usepackage{algorithm}
\usepackage{algorithmicx}
\usepackage{algpseudocode}
\usepackage{algpseudocode}
\usepackage{graphicx}
\usepackage{subcaption}
\usepackage{lipsum} 
\usepackage{amsmath}          
\usepackage{amssymb}          
\usepackage{siunitx}
\usepackage{threeparttable}
\usepackage{xcolor}
\newcommand{\gain}[1]{\textbf{($\uparrow$#1)}}
\usepackage{makecell}
\usepackage{booktabs}
\usepackage{hyperref} 
\usepackage{cleveref} 
\usepackage{fontawesome5}

\newcommand{\githubc}[1]{%
  \href{#1}{\faGithub\ #1}%
}

\def\BibTeX{{\rm B\kern-.05em{\sc i\kern-.025em b}\kern-.08em
    T\kern-.1667em\lower.7ex\hbox{E}\kern-.125emX}}
\begin{document}

\title{ReactionTeam: Teaming Experts for Divergent Thinking Beyond Typical Reaction Patterns 
}


\author{Taicheng Guo$^1$,
  \hspace{1pt} 
Changsheng Ma$^3$, \hspace{1pt}
Xiuying Chen$^3$, \hspace{1pt} 
Bozhao Nan$^2$, \hspace{1pt}
Kehan Guo$^1$ \\
Shichao Pei$^1$, \hspace{1pt} 
Olaf Wiest$^2$, \hspace{1pt}
Nitesh V. Chawla$^1$, \hspace{1pt} 
Xiangliang Zhang$^1$*\thanks{*Corresponding author.} \\

$^1$Department of Computer Science and Engineering, University of Notre Dame, \\
$^2$Department of Chemistry and Biochemistry, University of Notre Dame \\
$^3$King Abdullah University of Science and Technology\\
\{tguo2, xzhang33\}@nd.edu \\[5pt]
\githubc{https://github.com/taichengguo/ReactionTeam}\\
}

\maketitle

\begin{abstract}
Reaction prediction,  a critical task in synthetic chemistry, is to predict the outcome of a reaction based on given reactants. 
Generative models like Transformer have typically been employed to predict the reaction product. 
However, these likelihood-maximization models overlooked the inherent stochastic nature of chemical reactions, such as the multiple ways electrons can be redistributed among atoms during the reaction process. In scenarios where similar reactants could follow different electron redistribution patterns, these models typically predict the most common outcomes, neglecting less frequent but potentially crucial reaction patterns. These overlooked patterns, though rare, can lead to innovative methods for designing synthetic routes and significantly advance synthesis techniques.
To address these limitations, we build a team of expert models to capture diverse plausible reaction outcomes for the same reactants, mimicking the divergent thinking of chemists. The proposed framework, ReactionTeam, is composed of specialized expert models, each trained to capture a distinct type of electron redistribution pattern in reaction, and a ranking expert that evaluates and orders the generated predictions.
Experimental results across two widely used datasets and different data settings demonstrate that our proposed method achieves significantly better performance compared to existing state-of-the-art approaches.
\end{abstract}


\input{1-Introduction-new}
\input{2-related_work}

\input{3-Methodology}
\input{4-Experiments}

\input{5-Conclusion}

\bibliographystyle{IEEEtran}
\bibliography{sample-base}

\input{6-Appendix}

\end{document}

%% file: 1-Introduction-new.tex
\section{Introduction}

Reaction outcome prediction is one of the fundamental problems in computer-assisted organic synthesis. The aim of this task is to predict the products formed given a set of reactants and reagents~\cite{corey1969computer,coley2017prediction,keto2024data}.
Prior solutions for this task can be grouped into two main categories: template-based~\cite{segler2017modelling,segler2017neural} and template-free~\cite{jin2017predicting}. Template-based methods are typically rule-based and rely heavily on expert knowledge for generalizing reaction patterns and mechanistic pathways. The  knowledge requirement and the labor-intensive nature of template creation constrain these methods’ scalability and adaptability, struggling with novel or unconventional reactions. 
To tackle these limitations, template-free methods leverage deep generative models to predict   products from reactants. 
Indeed, while  generative models such as  Seq2Seq~\cite{sutskever2014sequence, guo2023can} are powerful  for predicting chemical reaction products, their capability is limited by
the training objective: maximizing the likelihood of the correct output given a training dataset. 
As a result, common mapping patterns, those that appear frequently in the data,  are effectively captured, while  non-typical patterns, which may correspond to rare but chemically valid electron redistributions, are often overlooked. This limitation is particularly problematic in chemistry, where  multiple valid products can arise from the same set of reactants due to alternative electron redistribution pathways, reaction conditions, or catalysts.

\begin{figure}[t]
\centering
\includegraphics[width= 0.42\textwidth]{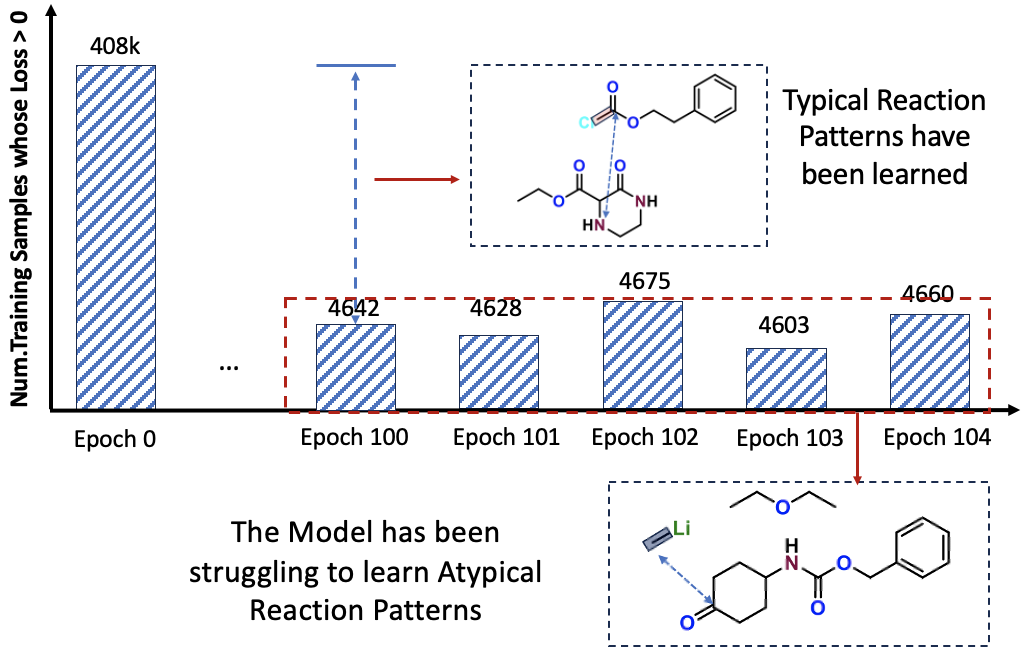} \vspace{-0.1in}
\caption{Number of training samples whose $loss > 0$ over training epochs.
} \vspace{-0.2in}
\label{fig:intro_problem} 
\end{figure}

A likelihood-based model tends to treat these less frequent outcomes as noise or anomalies, and thus systematically suppresses them during optimization. As shown in Fig. \ref{fig:intro_problem}, training loss plateaus after a certain number of epochs, indicating that the model saturates on the easily learnable, high-frequency patterns. However, a long tail of ``\textbf{hard-to-learn}'' examples remains. They correspond to \textbf{atypical reactions} that the model fails to predict accurately.

Crucially, many of these atypical reactions involve reactants that are structurally similar to those found in well-learned, typical reactions. This suggests that the model's failure may be  due to \textbf{underexploration of alternative} mechanistic possibilities. The log-likelihood objective, by \textbf{penalizing uncertainty}, discourages exploration of such alternatives, pushing the model toward overconfident, mode-seeking behavior.

\begin{figure}
\centering
\includegraphics[width= 0.48\textwidth]{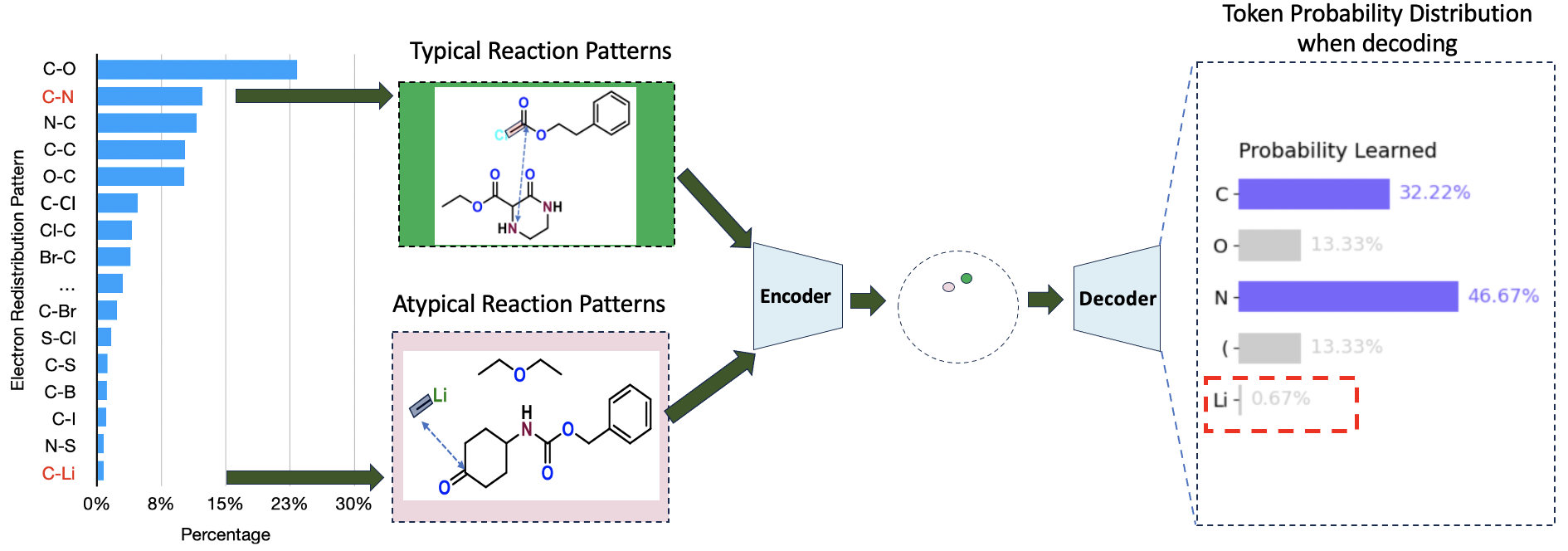}
\caption{
Reactants that follow typical or atypical electron redistribution patterns are encoded into similar representations due to structural similarity. The decoder learns to favor   high-probability tokens (e.g.,   \texttt{C} and  \texttt{N}),  while underrepresenting tokens necessary for atypical reactions (e.g.,  \texttt{Li}).
 } \vspace{-0.1in}
\label{fig:intro_problem2} 
\end{figure}

These limitations persist into inference time. Maximum-likelihood trained models, such as  Transformer-based encoders and decoders, generate outputs deterministically via greedy or beam search ~\cite{freitag2017beam}, further reinforcing high-probability, typical outcomes, as illustrated in Fig. \ref{fig:intro_problem2}. While introducing sampling temperature during decoding can inject randomness, this method uniformly perturbs token probabilities, often leading to chemically implausible products.

This contrasts sharply with how expert chemists approach synthesis. They routinely consider multiple plausible reaction mechanisms, including low-probability pathways that may lead to novel or more efficient synthetic strategies. A model constrained by likelihood maximization lacks this divergent reasoning ability, and thus misses valuable opportunities for discovery and innovation.

To overcome these limitations, we propose \textbf{ReactionTeam}, a novel framework that embraces the  divergent thinking paradigm by constructing a team of specialized expert models. While mixture-of-experts (MoE) architectures have been widely explored, they are typically designed to partition distinct inputs to different experts using a learned gating function, where each expert handles a different region of the input space \cite{chen2022towards}.  However, our goal in reaction prediction is to capture \textbf{multiple valid outputs} for the same input, reflecting the non-deterministic nature of chemical reactions.

To this end,  expert models in ReactionTeam share the majority of their parameters,  encoding the  common understanding of fundamental reaction behavior,  but introduce  minimal weight differences via lightweight adaptations (LoRA) to express diverse yet chemically plausible perspectives. This design reflects the intuition that chemists often agree on general reaction logic while differing in their views on rare or context-sensitive transformations. Furthermore, instead of training all experts in parallel, we adopt a \textbf{sequential training strategy}.
Concretely, a primary expert is first trained to capture the typical reaction patterns.
Samples that the primary model consistently mispredicts are then passed to a new expert (a LoRA module), trained to specialize in those challenging cases. This process is repeated, encouraging each expert to take responsibility for a distinct subset of atypical or underrepresented reaction patterns. 
This sequentially trained Mixture-of-LoRA Experts (Seq. MOLE) form a team of divergent experts, each specializing in a unique subset of reaction patterns across the full training set. 

During inference, all experts in our framework generate predictions from their own specialized perspectives,  typical or atypical, reflecting the diverse reasoning pathways chemists might consider. This is different from traditional Mixture-of-LoRA Experts models \cite{wu2024mixture},  which use a gating function to assign similar reactants to the same expert and often resulting in only the most likely prediction.  Our framework is akin to boosting-style ensemble methods, where later learners focus on the residual errors of earlier ones. However, rather than merely aggregating predictions to improve overall accuracy, we go further by explicitly encouraging diversity and structure in the output space. Specifically, we introduce two key enhancements during inference:  1) We encourage small-scale divergence within each expert by applying \textbf{inference-time dropout}, which introduces controlled stochasticity into the decoding process. This enforces subtle variations in predictions from the same expert, allowing it to explore multiple plausible product candidates even under a fixed input.   2) We employ a dedicated \textbf{ranking expert} to evaluate and prioritize the outputs from all experts. This model  assesses chemical plausibility and synthetic relevance, selecting the most viable products from the diverse candidate set.

Our main contributions are summarized as follows: 
\begin{itemize} [leftmargin=*]
    \item To the best of our knowledge, \textbf{we are the first} to investigate the issue of divergent reaction prediction.  
    This is a critical and specialized challenge presented to generative models when applied to scientific discovery where  multiple plausible outcomes must all be considered for their underlying mechanistic validity. However, this requirement fundamentally conflicts with the behavior of most  max-likelihood based AI models, which are inherently biased toward predicting the most frequent outcome in the data. 
    \item We innovatively propose to solve these challenges by employing a team of experts. We detail this insight by designing Seq. MOLE and inference-time dropout to inclusively capture diverse reaction patterns by different experts, and then apply ranking expert to rank these predicted products.
    \item  We validate the effectiveness and generality of our method not only on standard USPTO-MIT and USPTO-STEREO datasets~\cite{jin2017predicting} but also under varied data settings, demonstrating consistent improvements in Top-N accuracy for both atypical reactions and the overall test sets.
\end{itemize}

%% file: 2-related_work.tex
\section{Related Work} 
\textbf{Reaction Prediction:} Methods for reaction prediction can be categorized as template-based and template-free. Traditional Template-based methods~\cite{segler2017modelling,segler2017neural} rely heavily on expert knowledge and tend to be less generalizable. 
Current template-free methods model reactions as a sequence of graph edits~\cite{shi2020graph,coley2017prediction} and current most state-of-the-art methods employ the Transformer as the decoder to generate the product (in the form of SMILES string) given the reactants~\cite{liu2017retrosynthetic,schwaller2019molecular,guo2023can,fang2024moltc,10.1145/3583780.3614865}.  
\textit{Our framework can be applied to any transformer-based method. In this paper, we equip our framework on Graph2Smiles~\cite{tu2022permutation}, which achieves the best performance among all previous methods, to demonstrate the effectiveness of our approach.}

\vspace{+0.05in}
\noindent \textbf{Mixture-of-Experts and Mixture-of-Lora Experts:} 
MoE has long been studied in the machine learning community~\cite{jordan1994hierarchical,jacobs1991adaptive}, where multiple experts are trained to specialize on different parts of the input and their outputs are aggregated~\cite{chen2022towards}. Recent work on mixture of LoRA experts~\cite{wu2024mixture} extends this framework by applying multiple LoRAs at each layer.
\textit{However, the key challenge in reaction prediction is capturing the diverse reaction patterns,  which requires experts to specialize in learning distinct mappings from reactants to products,  rather than being solely assigned to handle input reactants based on their surface-level features.
Therefore, \textbf{unlike previous MoE or MoLE experts} methods that focus on routing the input (reactants) to different experts, we specially design a sequential MOLE training method that enables each expert to specialize in a unique subset of electron redistribution patterns. Collectively, the expert team  models a wide range of plausible reaction pathways.} 

\vspace{+0.05in}
\noindent \textbf{Divergent Thinking:} 
Divergent thinking is a fundamental aspect of creativity, characterized by the ability to generate multiple and varied responses to open-ended prompts~\cite{Thestructureofintellect.}. This concept is particularly well-suited for chemical reaction prediction, where a single set of reactants may lead to a range of plausible products depending on conditions, catalysts, or mechanistic pathways. By modeling a spectrum of plausible outcomes, divergent thinking more accurately mirrors how chemists reason multiple plausible products when extending beyond typical reaction patterns. \textit{To our knowledge, \textbf{this is the first work to formally introduce divergent thinking into chemical reaction modeling}, analyze specific types of divergence relevant to chemical reasoning, and design solutions to encourage reaction-level ideational diversity.}


%% file: 3-Methodology.tex
\section{The Proposed ReactionTeam Framework} 

%


\subsection{Preliminary: 
a Single Expert for Reaction Prediction}
\label{sec:single-expert}
Following the definition of previous reaction prediction work~\cite{tu2022permutation}, 
we first introduce the Encoder-Decoder architecture of a single expert for reaction prediction.
The reaction prediction task can be formulated as a transformation from reactants $G^r = (V^r, E^r)$ to product SMILES $p = \{s_1, s_2, \ldots, s_n\}$, where $G^r$ denotes the molecular graph of reactants, $V^r$ denotes the atoms, and $E^r$ denotes the number of shared electrons between atoms in the reactants and products. The input molecular graph is constructed from their SMILES strings using   RDKit~\cite{RDKit}, which parses the SMILES and accurately reconstructs the molecular structure, including aromaticity, valency, and bond types. Such graph-based inputs can better capture the  underlying molecular structure  and connectivity, and facilitate the reaction outcome prediction. The output product  $p$ is a sequence of   tokens 
$\{s_1, s_2, \ldots, s_n\}$ which are obtained from canonical SMILES using the tokenizer in~\cite{schwaller2019molecular}. 

Given reactants graph $G^r$, an encoder is applied to transform  reactants as representation vectors. The Encoder could be implemented by a Directed Graph Convolutional Network (D-GCN)~\cite{somnath2021learning, yang2019analyzing} and followed by a Transformer encoder~\cite{bahdanau2014neural, vaswani2017attention} to obtain atom-level hidden representations $h^r$:
\begin{equation}
h^r = \text{Transformer-Encoder}(\text{D-GCN}(G^r)).
\end{equation}

Then a Transformer decoder is applied to decode from the atom representations $h^r$ to product sequences $p$:
\begin{equation}
p = \{s_1, s_2, \ldots, s_n\} =  \text{Transformer-Decoder}(h^r).
\end{equation}

The training objective is to learn a model $f$ that maximizes the conditional likelihood of the output product SMILES sequence given the input reactants graph:
\begin{equation}
\max_f\; \log f(p \mid G^r) = \sum_{i=1}^{n} \log f(s_i \mid s_1, \dots, s_{i-1}, G^r),
\end{equation}
where $f$ denotes the full graph-to-sequence model (D-GCN, Transformer encoder and decoder).
Note that the training set is typically composed of paired samples $\{G^r,p\}$. During inference, the model  $f$  predicts a ranked list of potential products,  ordered by their  predicted likelihood.  

\subsection{Sequential Mixture-of-LoRA Experts (Seq. MOLE)}

\subsubsection{The Training Pipeline}

\begin{figure*}[ht]
\centering  
\includegraphics[width= 0.78\textwidth]{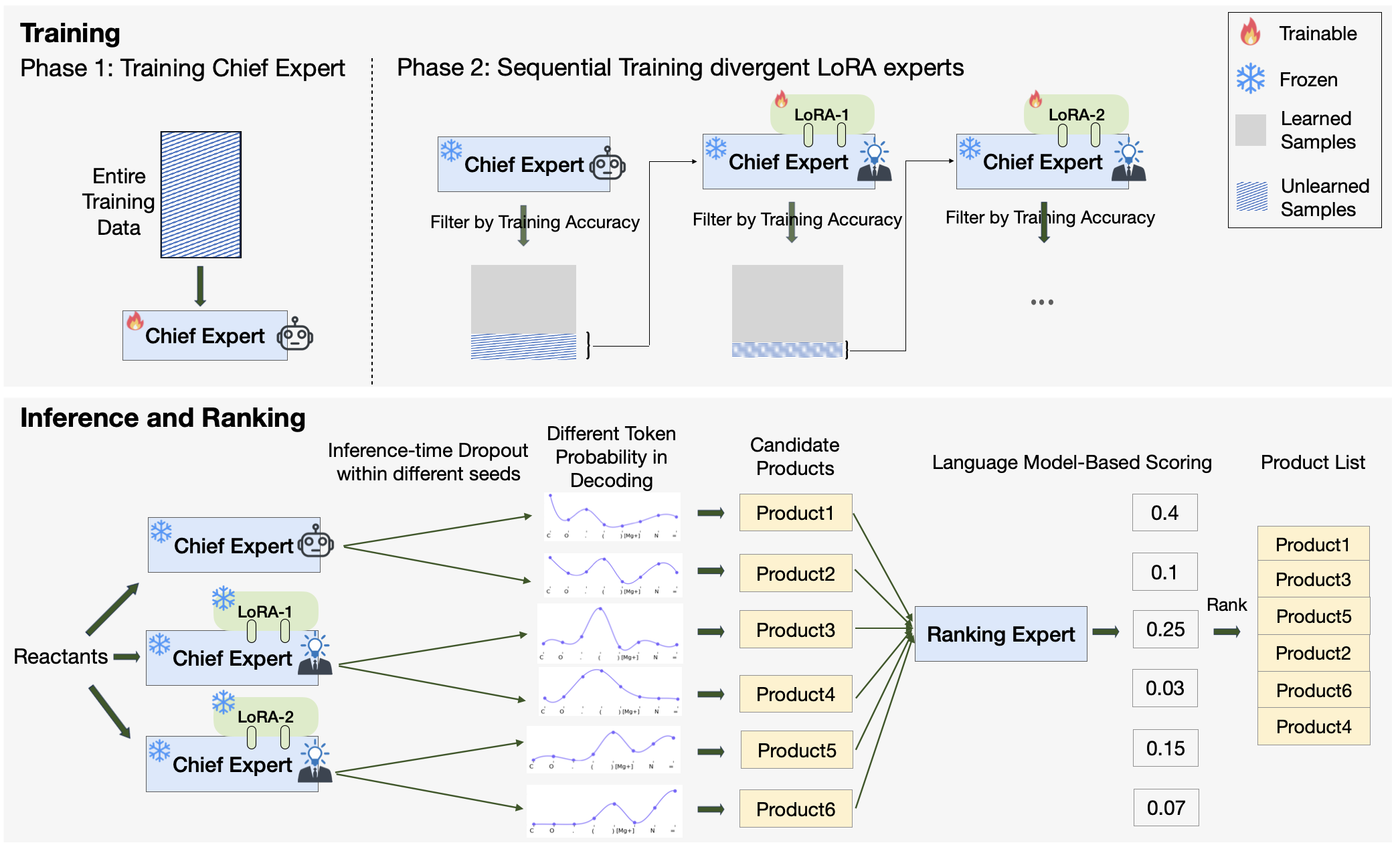}
\caption{The overall framework of our proposed solution ReactionTeam. } 
\vspace{-0.2in}
\label{fig:framework} 
\end{figure*}

To encourage substantial divergence among experts, we adopt a sequential training strategy, as illustrated in Fig.~\ref{fig:framework}. 
The pipeline begins by training a chief model $f_{\text{chief}}$ (e.g., a single expert model as described in  subsection \ref{sec:single-expert}) on the entail training dataset to capture the dominant and typical reaction patterns.  As shown in Fig. 1, the training loss plateaus
after capturing these common patterns are learned, indicating that the model struggles to fit the remaining \textit{hard-to-learn} cases.


Then, a set of LoRA-based  expert models $F$ = $\{f_1, f_2, \dots, f_n\}$ is trained sequentially, each focusing on the subset of training samples that were consistently mispredicted by the previous experts. The core idea is that the reaction patterns learned by a model are directly reflected in its ability to predict the correct product sequences. If a model is well-suited to a certain type of transformation, it will achieve high training accuracy on corresponding samples. Conversely, if a model struggles with certain patterns, it will produce incorrect outputs on samples exhibiting those patterns. Therefore, the training accuracy on each example serves as a proxy for evaluating how well a model captures specific reaction mechanisms.

Based on this intuition, we partition the training data among experts by leveraging sample-level accuracy during training. Specifically, at the final epoch\footnote{\textcolor{black}{The final epoch is determined when the training loss no longer decreases.}} of training $f_{\text{chief}}$, we collect the set of samples that it fails to predict, denoted as $\mathcal{D}_1$:
\begin{equation}
   \mathcal{D}_1 = \{(G^r, p) \in \mathcal{D} \mid f_{\text{chief}}(G^r) \neq p  \}. 
\end{equation}

This set of samples are then used to train a divergent expert $f_1$, which \textcolor{black}{inherits the architecture and weights of the chief model $f_{\text{chief}}$, keeps all base parameters frozen, and applies LoRA adaptation to introduce lightweight, trainable modules. } This allows $f_1$ to specialize in the mispredicted cases without interfering with the core knowledge captured by the chief model, promoting targeted divergence with minimal parameter overhead.
During the final epoch of training $f_1$, we similarly record the set of training samples that $f_1$ fails to predict correctly, denoted as $\mathcal{D}_2$:

\begin{equation}
    \mathcal{D}_2 = \{(G^r, p) \in \mathcal{D}_1 \mid    f_{1}(G^r) \neq p  \}. 
\end{equation}

We then save the LoRA weights of $f_1$ and proceed to train the next expert $f_2$ on $\mathcal{D}_2$, using $f_1$ as the initialization. The base model $f_{\text{chief}}$ remains frozen throughout, and only the LoRA parameters are updated. This process is repeated iteratively: at each step, the current expert is trained on the remaining subset of examples not yet well handled, and new experts are created until either the dataset is fully partitioned or a predefined number of experts is reached. For simplicity, each expert $f_i$ is trained over one iteration.

This sequential training scheme effectively partitions the dataset based on sample-level prediction accuracy. As a result, each expert $f_i$ is trained on a distinct and progressively more challenging subset of the data, potentially representing conflicting or underrepresented reaction patterns. This encourages each expert to learn from different reaction patterns and thus acquire a unique perspective on the reaction landscape.

Finally, we obtain the frozen base model $f_{\text{chief}}$ alongside a set of multi-perspective LoRA experts $F = \{f_1, f_2, \dots, f_n\}$. While $f_{\text{chief}}$ captures the dominant, typical reaction patterns, the experts in $F$ complement it by specializing in divergent and complex reaction patterns. Together, this ensemble enhances the system’s capacity to generalize across the reaction space while catching on both typical and atypical reaction patterns.

\subsubsection{LoRA Experts Configuration}
As discussed above, during the training of LoRA experts, we freeze all parameters of the base model $f_{\text{chief}}$ and apply LoRA adaptation. This design choice is motivated by two key considerations:

\begin{itemize}[leftmargin=*]
    \item \textbf{Effectiveness:} The LoRA experts are specifically intended to capture divergent reaction patterns. Since the datasets assigned to these experts are significantly smaller and more specialized than the full training set used for $f_{\text{chief}}$, a limited number of trainable parameters is sufficient to model the underlying transformations.
    \item \textbf{Efficiency:} Our framework maintains a set of diverse expert models, with LoRA adaptation reducing the memory and computation needed to train and store them, making the system scalable.
\end{itemize}

For simplicity and implementation efficiency, we adopt standard LoRA adaptation on each attention layer within the Transformer decoder. Specifically, we apply LoRA updates to the query (Q), key (K), and value (V) projections prior to their use in the attention mechanism. This setup enables each expert to learn distinct reaction behaviors with minimal parameter overhead while preserving the foundational knowledge encoded in $f_{\text{chief}}$.

{\footnotesize
\begin{equation}
\begin{aligned}
\tilde{Q} &= Q + \Delta Q 
          = \left(I + \frac{\alpha}{r} B_q A_q\right) Q, \\
\tilde{K} &= K + \Delta K 
          = \left(I + \frac{\alpha}{r} B_k A_k\right) K, \\
\tilde{V} &= V + \Delta V 
          = \left(I + \frac{\alpha}{r} B_v A_v\right) V.
\end{aligned}
\end{equation}
}

Here, $A_q, A_k, A_v \in \mathbb{R}^{r \times d}$ and $B_q, B_k, B_v \in \mathbb{R}^{d \times r}$ are the trainable low-rank matrices used to capture specialized reaction patterns in the query, key, and value projections, respectively. The hidden dimension is denoted by $d$, and $r \ll d$ is the LoRA rank. The scaling factor $\alpha$ controls the contribution of the LoRA update, and the ratio $\frac{\alpha}{r}$ determines its effective scale.
To ensure that the LoRA-augmented model initially behaves identically to the pre-trained backbone, we initialize the low-rank matrices with $A$ drawn from a random uniform distribution and $B$ initialized to zero.

By leveraging LoRA adaptation, we are able to construct a set of effective and divergent experts, each requiring only a small number of trainable parameters. The efficiency and effectiveness of this approach are quantitatively demonstrated in the experimental results provided in Section~\ref{sec:exp_mole}.

\subsection{Inference-time Dropout}
To introduce controlled stochasticity into the decoding process, we  explore dropout for creating small-scale divergence in each expert.
An analogy of this approach (training large-scale divergent experts and enabling small-scale divergence at inference) can be drawn from chess move prediction. A capable model should be able to predict actions for both common and rare board states (large-scale divergence), and it should also reflect the variability in human strategy (small-scale divergence) due to individual differences, emotional states, or situational nuances. In chemistry, similar variability arises from changes in reaction conditions such as temperature, solvent, or operator choice. We form a research question (RQ3) in Section \ref{Exp} to evaluate these two divergent thinking mechanisms. \textcolor{black}{The results presented in section \ref{sec:RQ3} verify that these two mechanisms complement each other.}

By applying dropout at test time and running multiple forward passes with different dropout masks, our insight is 
to randomly deactivate model weights, thereby generating lightweight variants of each expert. These variants serve as low-divergence models, each reflecting a different but minor interpretation of electron flow. Each model varied in this way can be viewed as a different \emph{expert} operating within a narrow space of plausible mechanistic alternatives.

This technique, known as Monte Carlo Dropout \cite{gal2016dropout}, aligns well with our objective of capturing the inherent stochastic nature of chemical reactions.
During the inference stage, we apply dropout to the full graph-to-sequence model (D-GCN, Transformer encoder and decoder). Since the dropout layers are influenced by the random seed, we vary the seed for each inference iteration to introduce variability in the predictions. For the chief expert $f_{chief}$ and each LoRA expert $f_{i}$, the inference-time dropout is repeated five times with different seeds to create different model weights for  \emph{candidate sampling} (introduced next). To obtain models with minor variations while retaining general chemical reaction patterns, a low dropout rate (0.1) is applied in our experiments.

\subsection{ Ranking During Inference}
\label{sec:gating}  

\subsubsection{Candidates Sampling}
For each model ($f_{chief}$ or $f_{i}$ each  with differently seeded dropout), we apply candidate sampling to generate predicted products. Specifically, we use beam search following  previous methods~\cite{tu2022permutation}. For one given  reactant, five candidate products are sampled from each expert, resulting a list of diverse candidate products $P = \{p_1, p_2, \dots, p_n\}$. 

\subsubsection{ Ranking Expert} 
The list of candidate products need to be ranked based on the likelihood that each would be produced from the given reactants. Since product generation is modeled as a token-by-token sequence prediction process, the likelihood of a candidate product $p = (s_1, s_2, \ldots, s_n)$ can be quantified by its joint probability under the model:

{\footnotesize
\begin{equation}
    Score(p \mid G^r) = \sum_{i=1}^{n} \log f_{\text{score}}(s_i \mid s_1, s_2, \ldots, s_{i-1}, G^r).
\end{equation}
}

Since it is impractical to retrain a separate scoring model, we choose to reuse the chief model for scoring candidate products for several reasons. First, although the chief model is not designed to generate diverse outputs, it serves well as a reliable evaluator.  It may operate within the dominant reaction space, but it has been trained on the entire training dataset, enabling it to judge how well a candidate product aligns with learned reaction patterns. In this sense, it acts as a strong ``examiner'' even if not a divergent ``thinker.'' Second, training a separate scoring model (such as a discriminative reranker) would require additional labeled data, which is often expensive or unavailable in chemistry. Third, our extensive evaluation (section \ref{sec:RQ2}) demonstrates that this reuse strategy is both effective and efficient, achieving strong ranking performance without introducing additional training complexity.

In the end, product candidates are ranked based on their scores. The top-ranked candidates represent the model's most confident and chemically plausible predictions. 

%% file: 4-Experiments.tex
\section{Experiments}
\label{Exp}

We conducted extensive experiments to address the following research questions (RQs):
\begin{itemize}[leftmargin=*, itemsep=2pt, parsep=0pt]
    \item \textbf{RQ1: Effectiveness and Generalization}. Can \textit{ReactionTeam} effectively encourage divergent thinking and consistently improve reaction prediction performance across different settings?
    \item \textbf{RQ2: In-Depth Analysis of Divergent-Thinking Mechanisms}. How do the divergent-thinking Mechanisms within ReactionTeam contribute individually and collectively?
    \item \textbf{RQ3: Expert Behavior and Scoring}. How do divergent LoRA experts behave during decoding, and why is the Chief Expert reasonable for scoring and ranking, even if it fails to generate atypical reaction patterns?
\end{itemize}

\subsection{Experiment Settings}
\textbf{Datasets and preprocessing.}
Same as most previous work~\cite{coley2019graph,schwaller2019molecular}, we mainly evaluate our method on the current widely public reaction prediction dataset USPTO-MIT~\cite{jin2017predicting}. There are 480K reactions in this dataset, and the default train-test splits are widely adopted in all baselines and our method. 
To assess our model's performance in capturing the ``atypical'' electron redistribution patterns, we specifically select reactions where the electron flow percentage is less than   1\%   and test our model on them. We also evaluate our methods on the full standard test set of USPTO-MIT to demonstrate their strong performance. To further validate the generalizability of our approach, we evaluate on USPTO-STEREO~\cite{uspto_stereo_dataset}, which contains 900K training and 50K test reactions with types distinct from USPTO-MIT. In addition, since no other large-scale datasets are publicly available, we apply random shuffling and splitting on USPTO-MIT, showing that our method consistently outperforms across different data distributions.


\vspace{+0.1in}
\noindent \textbf{Implementation details.} 
We implement our model using Pytorch 
and conduct all the experiments on a Linux server with GPUs (4 Nvidia A100). 
As mentioned in the previous methodology, we reuse the encoder and decoder architectures from the current state-of-the-art Graph2Smiles model. To ensure a fair comparison to previous autoregressive methods, we follow the same detailed settings as the previous model. 
For D-GCN, the number of message updating steps is set to 4. Following Molecular Transformer and Graph2Smiles, we fix the embedding and hidden sizes $d_{model}$ to 256, the filter size for the Transformer to 2048, the number of attention heads to 8, and the number of layers
for the attention encoder and the Transformer decoder both to 6. 
We use the AdamW optimizer~\cite{loshchilov2017decoupled} with Noam learning rate scheduler~\cite{vaswani2017attention}.
For the sequential training, we set the maximum sequential training iteration to 150. 
For the LoRA settings, we set the rank to 64, alpha to 32. 
For inference-time dropout, we randomly generate 10 dropout seeds and set a dropout out rate to 0.1.
For each expert sampling, we select the top 5 products of the expert and apply the beam search size of 50, which is the same as the  Graph2Smiles method.
As shown in the Section~\ref{sec:exp_mole}, we find that although we obtain multiple experts from sequential training, the incremental performance remains unchanged after adding 3 experts, so actually we only use three experts (one chief expert and two specialized experts) to predict the products. All implementation code and model checkpoints to reproduce the results will be made publicly available upon acceptance of the paper.

\vspace{+0.1in}
\noindent\textbf{Evaluation Metrics.}
Following prior work~\cite{coley2019graph, schwaller2019molecular, bi2021non, meng2023doubly}, we evaluate our model and baselines using Top-K accuracy, which measures the percentage of reactions where the ground-truth product appears among the top $K$ predictions.
It is important to note that our \textit{ReactionTeam} framework is explicitly designed to promote divergent thinking and generate a broader set of plausible reaction outcomes, including those beyond typical patterns. Therefore, we aim to demonstrate that our method increases the coverage of chemically reasonable candidates. To reflect this goal, we evaluate Top-K accuracy at various values of $K \in \{1, 2, 5, 10, 20, 50\}$, which allows us to assess how well our method expands the reaction prediction space through ideational diversity. Note that we mainly focus on Top-20 and Top-50 accuracy, which are critical in real pipelines (e.g., synthesis planning, virtual screening, safety/IP checks), because the model only needs to place the true product in a small shortlist that downstream filters or chemists can evaluate, so higher top-K accuracy directly translates to higher end-to-end success in practical applications.

\vspace{+0.1in}
\noindent\textbf{Baselines.}
We compare our method with following SOTA baselines:
\noindent\textbf{Autoregressive.} MT-based~\cite{schwaller2019molecular}, Chemformer~\cite{irwin2022chemformer}, Graph2Smiles~\cite{tu2022permutation} are various models for reaction prediction. MT-based treats the problem as SMILES string translation using transformers. Chemformer is also a transformer-based model and uses a pretrained SMILES encoder with self-supervised tasks. Graph2Smiles combines graph encoding with transformer decoding to achieve permutation-invariant encoding of molecule graphs. Our framework can be applied to any Transformer decoder-based method. We employ Graph2Smiles in our experiments since it's the current SOTA reaction prediction model. We  demonstrate that our framework can  work well and significantly boost performance for such strong model. 
\noindent\textbf{Non-autoregressive.} NERF~\cite{bi2021non} formulates reaction prediction as generating an electron redistribution matrix, producing its values in a non-autoregressive manner.

\subsection{RQ1: Effectiveness and Generalization} \vspace{-0.01in}

\begin{figure*}[t]
\centering
\begin{subfigure}[b]{0.3\textwidth}
    \includegraphics[width=0.9\textwidth]{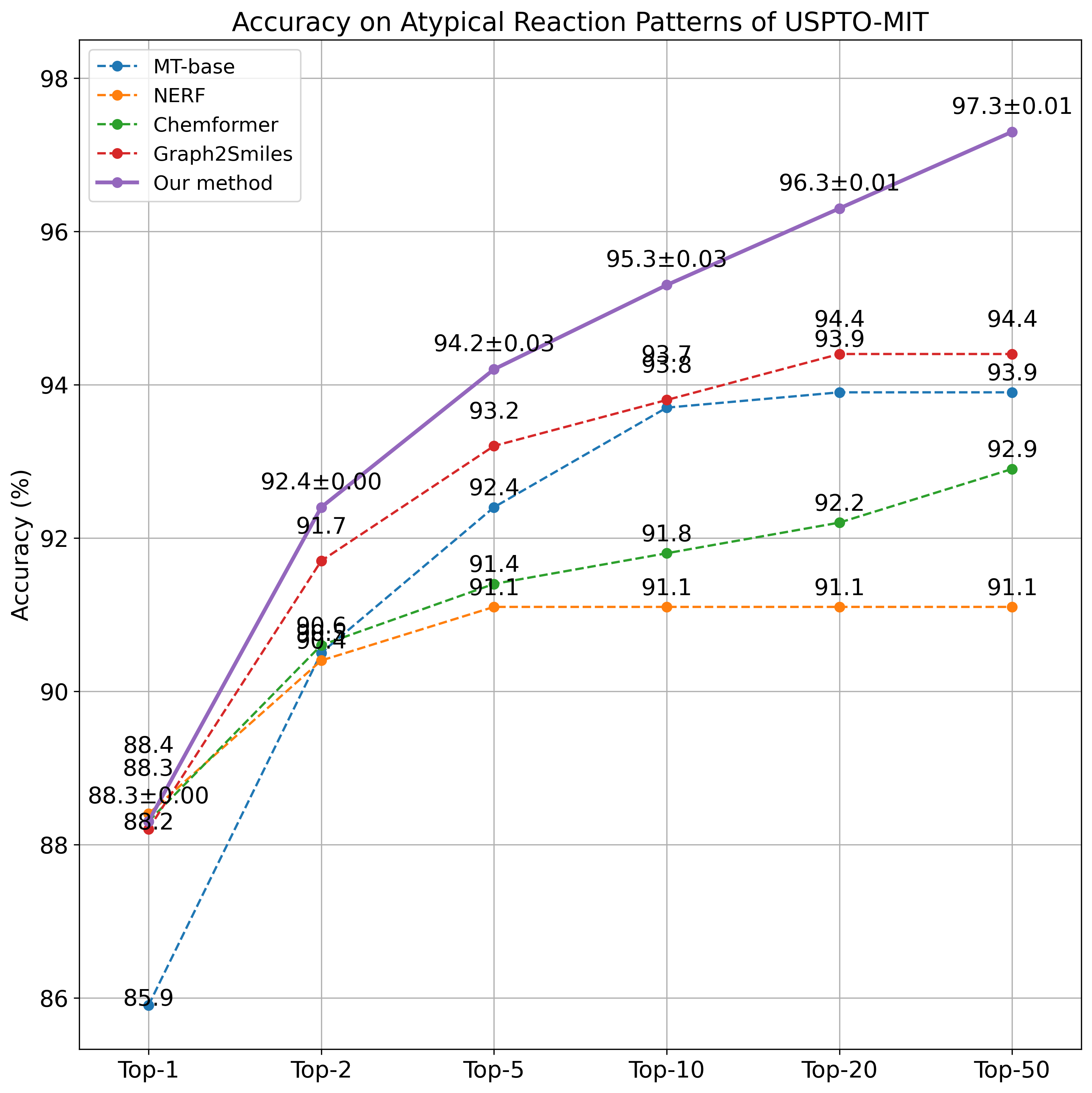}
    \caption{ USPTO-480K atypical reactions}
    \label{fig:intro_left}
\end{subfigure}
\hspace{0.3cm}
\begin{subfigure}[b]{0.3\textwidth}
    \includegraphics[width=0.9\textwidth]{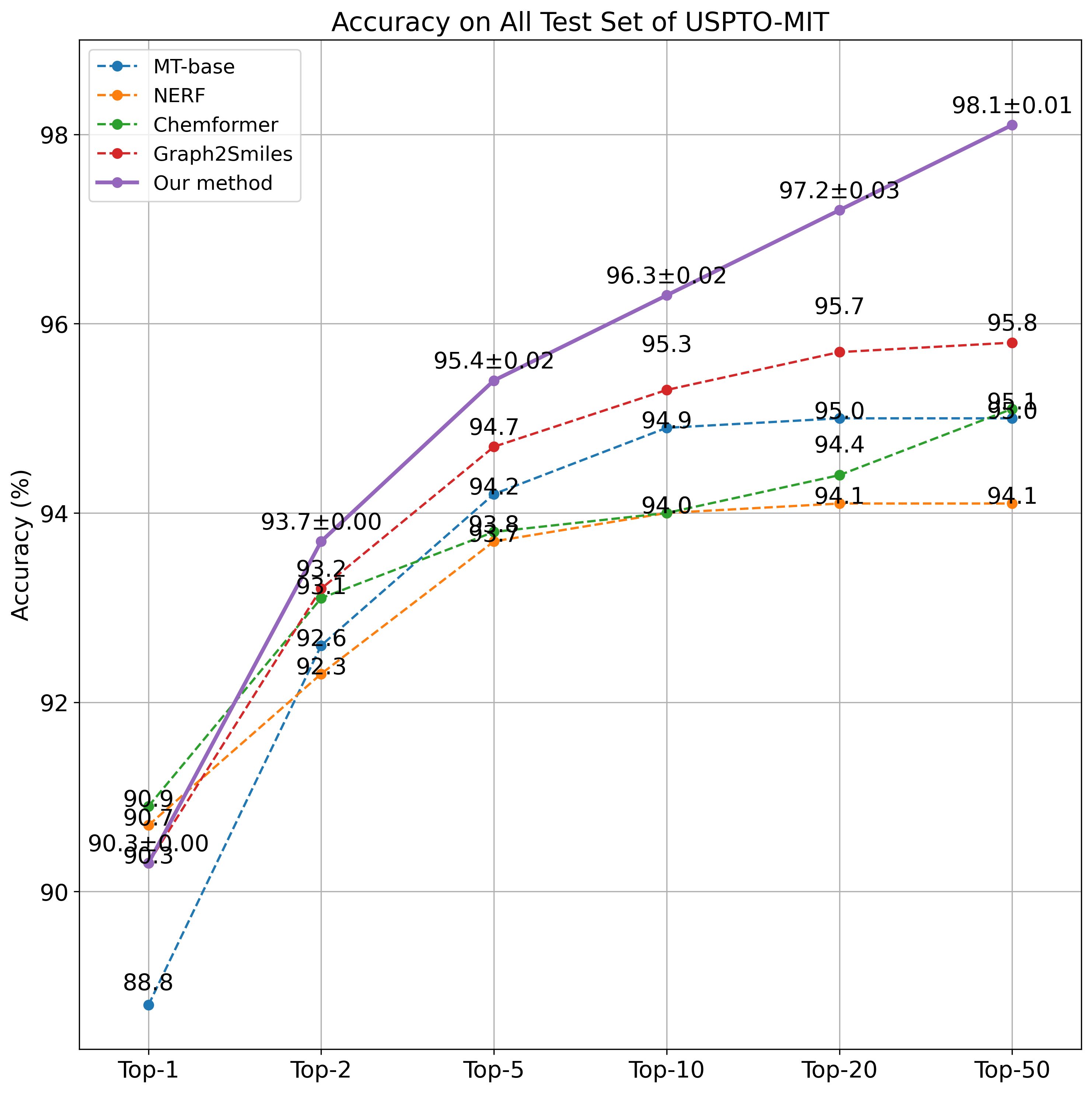}
    \caption{  USPTO-480K Standard test reactions} 
\end{subfigure}
\hspace{0.3cm}
\begin{subfigure}[b]{0.32\textwidth}
    \includegraphics[width=0.86\textwidth]{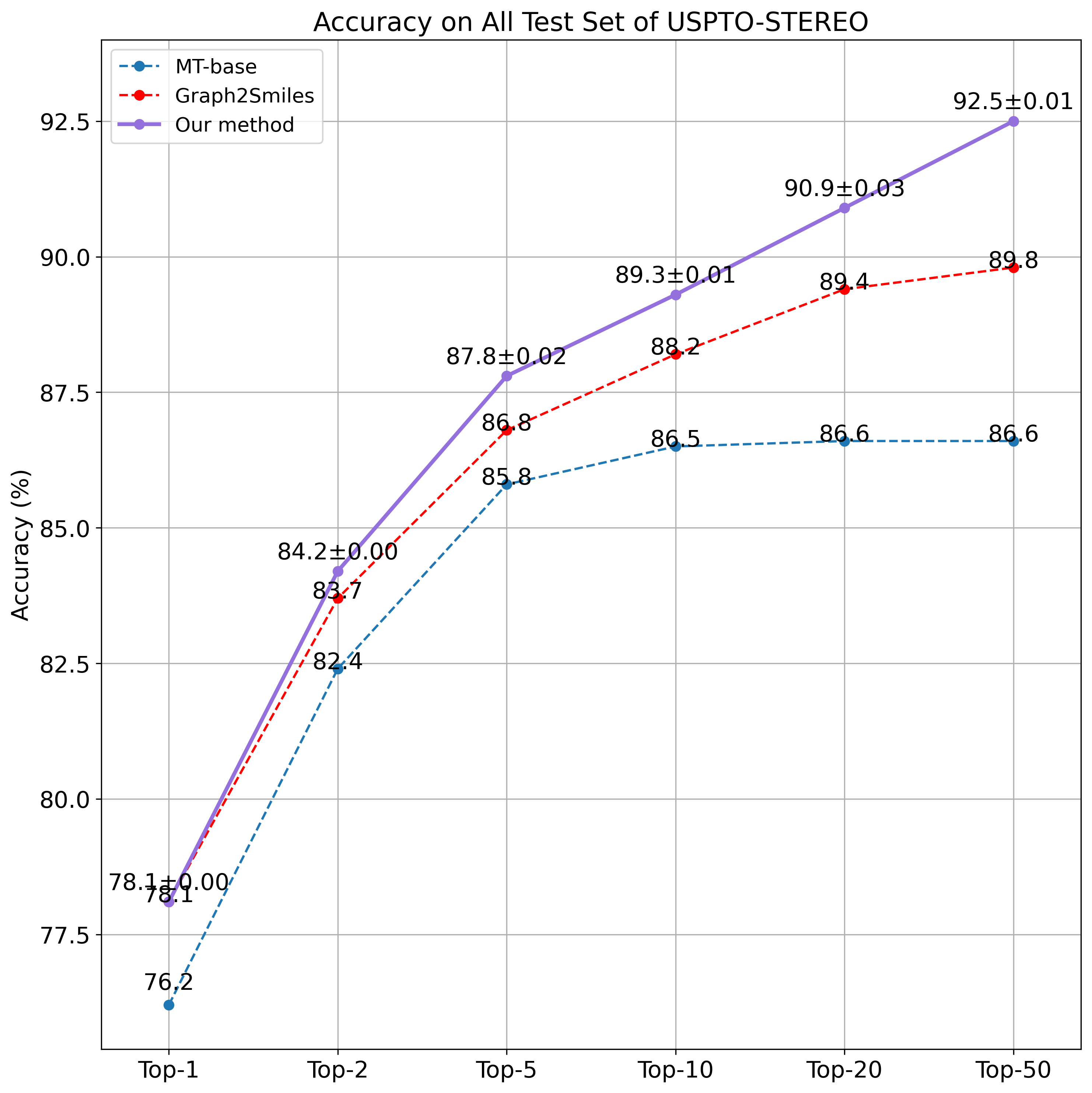}
    \caption{  USPTO-STEREO Standard test reactions} 
\end{subfigure}
\caption{The Top-K Accuracy (\%)  of evaluated models in the USPTO-MIT and USPTO-STEREO datasets.}
\label{fig:exp_all}
\vspace{-0.1in}
\end{figure*}

\subsubsection{RQ1.1 Does ReactionTeam achieve consistent improvements 
across different datasets and resamplings?}
\textcolor{black}{The performance of evaluated models for atypical reaction patterns and} overall performance measured by Top-K accuracy is reported in Fig.~\ref{fig:exp_all} and Table~\ref{Table:exp_different_shuffle}.

We can observe: \textbf{1) Improvements of prediction on atypical reaction patterns.}
We observe that our approach consistently outperforms all baselines across Top-2 to Top-50 accuracy, significantly improving upon the Graph2Smiles model and achieving the highest accuracy, from 92.4\% to 97.3\%, as shown in Fig.~\ref{fig:exp_all}(a). As noted in the Introduction, baseline methods show only marginal improvements as $k$ increases, suggesting that they struggle with atypical reactions and lack the flexibility to model diverse mechanistic pathways. In contrast, our approach exhibits a steadily increasing performance curve with larger $k$, indicating its ability to generate multiple plausible products, particularly for atypical cases. This highlights the importance of diversifying reaction predictions to improve real-world reaction modeling. The observed gains can be attributed to the effectiveness of our divergent expert models and the ranking mechanism. \textbf{2) Improvements in prediction on all reactions across different datasets and distributions.}
Our approach also significantly outperforms all previous methods on both the full USPTO-MIT and full USPTO-STEREO test set (in Fig.~\ref{fig:exp_all}(b) and (c)), which are two widely used reaction prediction benchmarks. The improvement trend mirrors that observed for atypical reaction patterns. Also, as shown in Table~\ref{Table:exp_different_shuffle}, we can observe that our method consistently outperforms the best baseline across different settings (different train: test ratio and shuffle). All these experiments highlight the strong generalization ability of our approach and its effectiveness across the full spectrum of reaction types and different datasets and distributions.
\textbf{3) Top-1 Accuracy.} 
We also report Top-1 accuracy in all experiments. Although our method is designed to encourage divergent thinking and improve diversity (Top-2 to Top-50), it does not compromise Top-1 performance. In fact, it matches or is better than the baseline, showing that diversity gains come without sacrificing precision on the most likely prediction.

\subsubsection{RQ1.2 Does ReactionTeam generate predicted products with greater diversity and novelty compared to baselines?}

\begin{figure} 
\centering
\includegraphics[width= 0.4\textwidth]{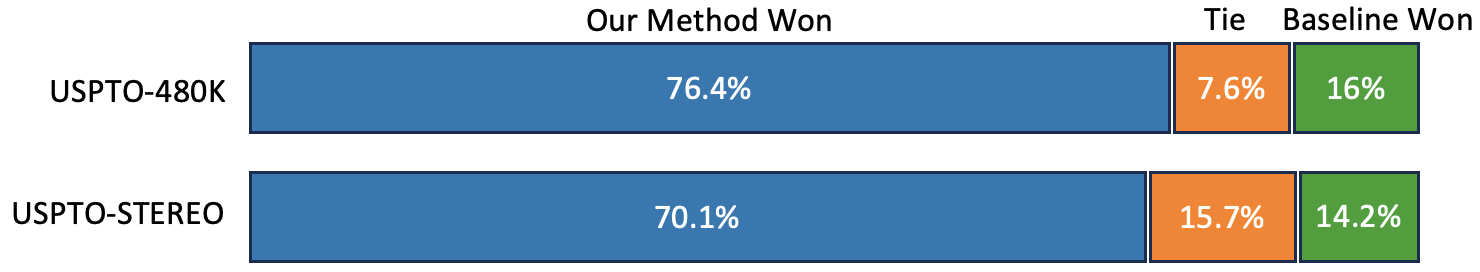}
\caption{Our Method V.S. Previous Best Baseline on predicted candidate products novelty and diversity, as judged by GPT-5.} 
\label{fig:exp_novel} 
\vspace{-0.1in}
\end{figure}

To evaluate diversity and novelty, we conduct two experiments. \textbf{(1) Diversity:} As shown in Table~\ref{Table:exp_different_shuffle}, we report the average internal diversity (Avg. ID) of predicted products, calculated as the mean Tanimoto similarity across all pairs of candidates per method. Our method consistently achieves higher Avg. ID under different settings, demonstrating improved diversity in product predictions. \textbf{(2) Novelty:} Given the 40K test samples, it is difficult to directly measure novelty. We therefore employ an LLM-as-judge strategy: GPT-5 is provided with the predictions from our method and from the best baseline, and asked to decide which set contains more novel and correct products. Each comparison is repeated three times, with averaged results shown in Fig.~\ref{fig:exp_novel}. Across both datasets, our method outperforms the strongest baseline, indicating its potential to generate novel and valid products. We note this is an auxiliary demonstration; the more critical evidence lies in our earlier results showing improved performance on atypical reactions.

\begin{table}[]
\centering
\caption{Performance on USPTO-480K under different data splits and shuffles. The ``Split Variant" reduces the train–test ratio, while the ``Shuffle Variant" randomizes USPTO-480K while keeping the original ratio. 
 Top-$k$ accuracy (\%) and average internal diversity (Avg. ID) are reported. }
\resizebox{0.5\textwidth}{!}{
\begin{tabular}{llllll}
\hline
\multicolumn{1}{c}{Setting} & \multicolumn{1}{c}{Method} & \multicolumn{1}{c}{Top-1$\uparrow$} & \multicolumn{1}{c}{Top-20$\uparrow$} & \multicolumn{1}{c}{Top-50$\uparrow$} & \multicolumn{1}{c}{Avg. ID$\uparrow$} \\ \hline
\multirow{2}{*}{\makecell[l]{Standard split \\ (Train:Test=10:1)}} 
  & Baseline & 90.7 & 95.7 & 95.8 & 0.48 \\
  & \cellcolor{gray!15}Ours     
         & \cellcolor{gray!15}{90.9} 
         & \cellcolor{gray!15}{97.2} 
         & \cellcolor{gray!15}{\makecell[l]{98.1 \hfill \scriptsize \gain{2.4\%}}} 
         & \cellcolor{gray!15}{\makecell[l]{0.56 \hfill \scriptsize \gain{17\%}}} \\ \hline
\multirow{2}{*}{Atypical subset} 
  & Baseline & 88.3 & 94.4 & 94.4 & 0.56 \\
  & \cellcolor{gray!15}Ours     
         & \cellcolor{gray!15}{88.4} 
         & \cellcolor{gray!15}{96.3} 
         & \cellcolor{gray!15}{\makecell[l]{97.3 \hfill \scriptsize \gain{3.1\%}}} 
         & \cellcolor{gray!15}{\makecell[l]{0.60 \hfill \scriptsize \gain{7\%}}} \\ \hline
\multirow{2}{*}{\makecell[l]{Split Variant-1 \\ (Train:Test=3.33:1)}} 
  & Baseline & 84.6 & 94.6 & 95.1 & 0.47 \\
  & \cellcolor{gray!15}Ours     
         & \cellcolor{gray!15}{84.6} 
         & \cellcolor{gray!15}{95.4} 
         & \cellcolor{gray!15}{\makecell[l]{96.4 \hfill \scriptsize \gain{1.4\%}}} 
         & \cellcolor{gray!15}{\makecell[l]{0.56 \hfill \scriptsize \gain{19\%}}} \\
\multirow{2}{*}{\makecell[l]{Split Variant-2 \\ (Train:Test=6.66:1)}} 
  & Baseline & 86.1 & 95.1 & 95.6 & 0.44 \\
  & \cellcolor{gray!15}Ours     
         & \cellcolor{gray!15}{86.1} 
         & \cellcolor{gray!15}{96.1} 
         & \cellcolor{gray!15}{\makecell[l]{97.2 \hfill \scriptsize \gain{1.7\%}}} 
         & \cellcolor{gray!15}{\makecell[l]{0.52 \hfill \scriptsize \gain{18\%}}} \\ \hline
\multirow{2}{*}{\makecell[l]{Shuffle Variant-1 \\ (Train:Test=10:1)}} 
  & Baseline & 90.3 & 96.0 & 96.2 & 0.44 \\
  & \cellcolor{gray!15}Ours     
         & \cellcolor{gray!15}{90.4} 
         & \cellcolor{gray!15}{97.4} 
         & \cellcolor{gray!15}{\makecell[l]{98.1 \hfill \scriptsize \gain{2.0\%}}} 
         & \cellcolor{gray!15}{\makecell[l]{0.51 \hfill \scriptsize \gain{16\%}}} \\
\multirow{2}{*}{\makecell[l]{Shuffle Variant-2 \\ (Train:Test=10:1)}} 
  & Baseline & 90.1 & 96.2 & 96.3 & 0.41 \\
  & \cellcolor{gray!15}Ours     
         & \cellcolor{gray!15}{90.3} 
         & \cellcolor{gray!15}{97.6} 
         & \cellcolor{gray!15}{\makecell[l]{98.3 \hfill \scriptsize \gain{2.1\%}}} 
         & \cellcolor{gray!15}{\makecell[l]{0.51 \hfill \scriptsize \gain{24\%}}} \\ \hline
\end{tabular}
}
\label{Table:exp_different_shuffle}
\end{table}

\subsection{RQ2: Analysis of Divergent-Thinking Mechanisms}

\subsubsection{RQ2.1   Individual contributions of Seq. MOLE and Inference-Time Dropout}

\label{sec:RQ3}
To further investigate the contribution of each component in our approach, we conduct ablation studies to isolate the effects of our designed two divergent thinking modules: \textit{Seq. MOLE} and \textit{Inference-Time Dropout}. The results of different variants are shown in Table~\ref{abstudy}.  For the Baseline+Seq. MOLE setting, we do not apply dropout; candidate products are generated solely using the chief expert and its associated LoRA experts. In contrast, for the Baseline + Dropout setting, we apply inference-time dropout with different random seeds exclusively on the chief expert to produce candidate products.

\begin{table}[t!]
        \centering
        \caption{The impact of Seq.MoLE, Inference-time Dropout, and their combination.}
        \vspace{-0.05in}
        \resizebox{0.49\textwidth}{!}{
        \begin{tabular}{lcccccc}
\cline{1-7}
Model Name                     & Top-2$\uparrow$ & Top-5$\uparrow$ & Top-10$\uparrow$ & Top-20$\uparrow$ & Top-50$\uparrow$ & Avg.$\uparrow$ \\ \cline{1-7}
Baseline (Graph2Smiles)        & 91.7  & 93.2  & 93.8  & 94.4  & 94.4  & 93.5 \\
Baseline+Seq. MOLE           & 92.2  & 93.9  & 94.7  & 96.0  & 96.2 \gain{1.8\%} & 94.6 \\
Baseline+Dropout             & 92.4  & 94.2  & 94.9  & 95.9  & 96.7 \gain{2.4\%} & 94.8 \\
Baseline+Seq. MOLE+Dropout & \textbf{92.4}  & \textbf{94.3}  & \textbf{95.3}  & \textbf{96.4}  & \textbf{97.3 \gain{3.1\%}} & \textbf{95.2} \\ \cline{1-7}
\end{tabular}}
        \vspace{-0.1in}
        \label{abstudy}
\end{table}

We observe that Baseline+Seq. MOLE consistently improves performance across all Top-K accuracy metrics, achieving an average accuracy of 94.6\% compared to 93.5\% for the baseline. Similarly,  Baseline+Dropout yields an average accuracy of 94.8\%, also showing consistent gains over the baseline. The combination of both strategies (Baseline+Seq. MOLE+Dropout) achieves the best overall performance, with an average accuracy of 95.2\%. This result demonstrates that each mechanism contributes positively and that the divergent candidate products generated by Seq. MOLE and Inference-Time Dropout are complementary. Together, they expand the reaction prediction space in a synergistic manner, resulting in more accurate and diverse product predictions.

\subsubsection{RQ2.2 How sensitive is ReactionTeam’s performance to the number of divergent experts and dropout trials?}

\paragraph{The impact of the number of Experts}
\label{sec:exp_mole}
When sequentially training 
$F = \{f_1,   \dots, f_n\}$, several valid questions 
arise: \emph{How many  LoRA experts would be enough? How much do they contribute to the overall performance? And do they introduce significant storage overhead?}
We address these questions by evaluating both the \textbf{effectiveness} and \textbf{efficiency} of the LoRA experts. First, we analyze the number of experts produced and quantify their individual and collective impact on model performance. Then, we examine their efficiency by measuring the storage cost and demonstrating how LoRA adaptation allows us to maintain a compact model footprint while still benefiting from expert-level specialization.

\begin{figure}[t]
\centering
\includegraphics[width= 0.45\textwidth]{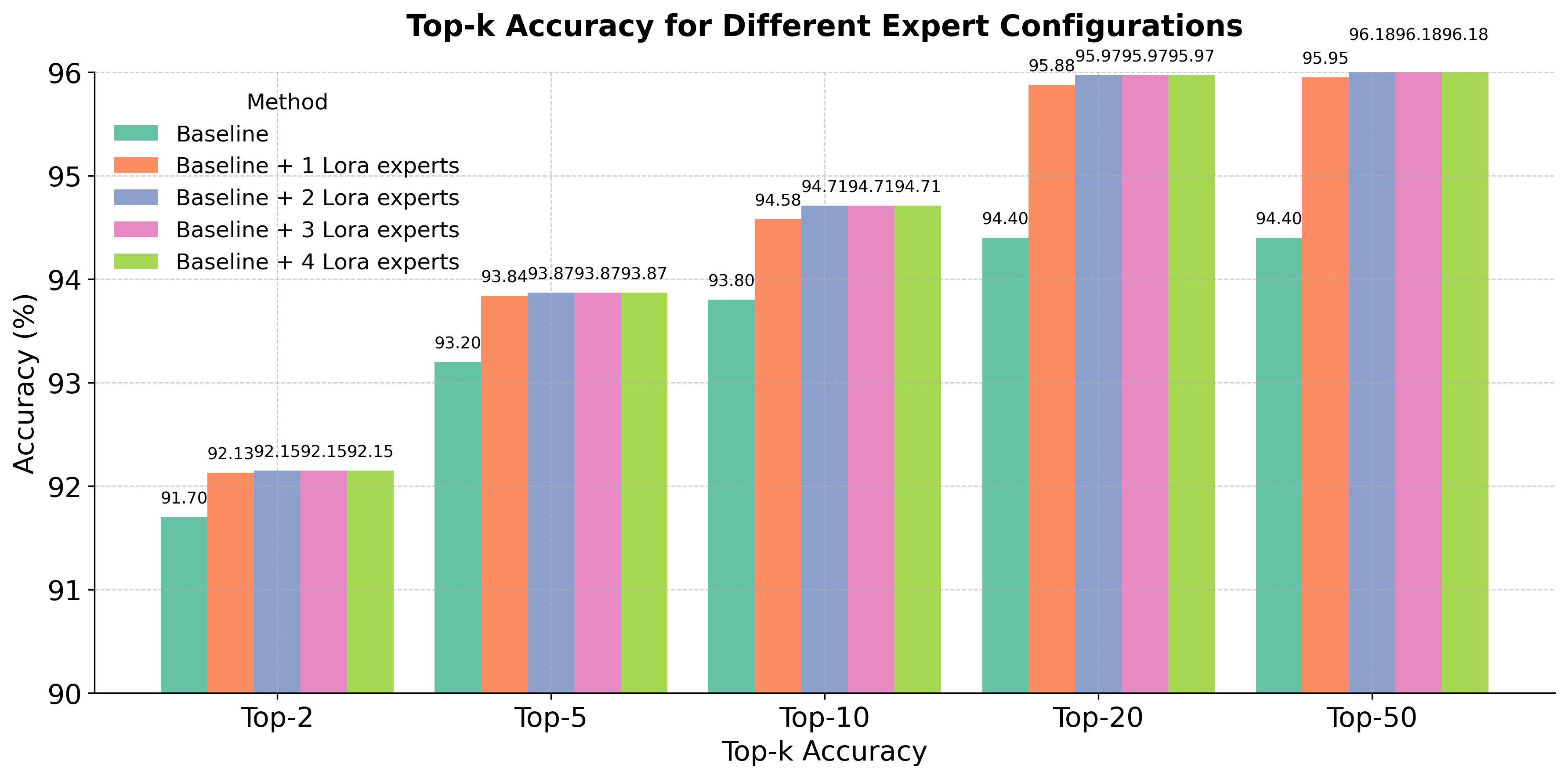}
\caption{Top-k Accuracy improvements with increasing number of LoRA experts. Adding the first two LoRA experts can introduce significant performance while no gain observed from adding the following LoRA experts.} 
\label{exp_LoRAexperts} 
\vspace{-0.15in}
\end{figure}

\noindent\textbf{The Effectiveness of MOLE.}
In Fig.~\ref{exp_LoRAexperts}, we show Top-k Accuracy when increasing LoRA experts. We observe:
\textbf{1)} There is a consistent increase in accuracy across all top-k metrics as more LoRA experts are added.
\textbf{2)} Adding even one LoRA expert gives a significant improvement. For instance, Top-50 accuracy jumps from 94.4\% to 95.95\%. Subsequent additions of LoRA experts (from 2 to 4) result in marginal improvements, with the performance plateauing from 2 experts onward. This is because filtering training samples using $f_{i}(G^r) \neq p $ results in most samples being assigned to the first two experts. Consequently, only a small portion of the training set remains for the remaining experts. Later LoRA experts thus contribute little and produce similar patterns. 
Hence, the first two LoRA experts are enough, and in our experiments, besides the chief expert, we use the additional two LoRA experts.  

\noindent\textbf{The Parameter Efficiency of MOLE.}
In Table~\ref{tab:LoRA_experts}, we show the total model parameters and the additional required parameters compared to the baseline. We can observe that although  two LoRA experts are involved, the number of additional parameters is small (only 1.18M) and  the performance gain is substantial  (around 2\% across all Top-k accuracy as shown in Fig.~\ref{exp_LoRAexperts}). This highlights the parameter efficiency of our approach alongside its improved performance.

\begin{table}[t!]
\centering
\caption{No. of  parameters, with and without LoRA experts.} \resizebox{0.42\textwidth}{!}{
\begin{tabular}{@{}lcc@{}}
\toprule
Method          & Total Params (M)         & \begin{tabular}[c]{@{}c@{}} Increase Compare   to  Baseline  (M) \end{tabular}      \\ \midrule
Baseline        & 18.31M (18,305,666)            & 0.00M                                            \\
Using Chief Expert    & 18.31M (18,305,666)            & 0.00M                                      \\
+1 LoRA Experts & 18.90M (18,895,490)            & +0.59M (589,824)                                   \\
+2 LoRA Experts & 19.49M (19,485,314)            & +1.18M (1,179,648)                                 \\ \bottomrule
\end{tabular}
}
\label{tab:LoRA_experts}
\vspace{-0.15in}
\end{table}

\begin{figure}[t]
\centering
\includegraphics[width= 0.4\textwidth]{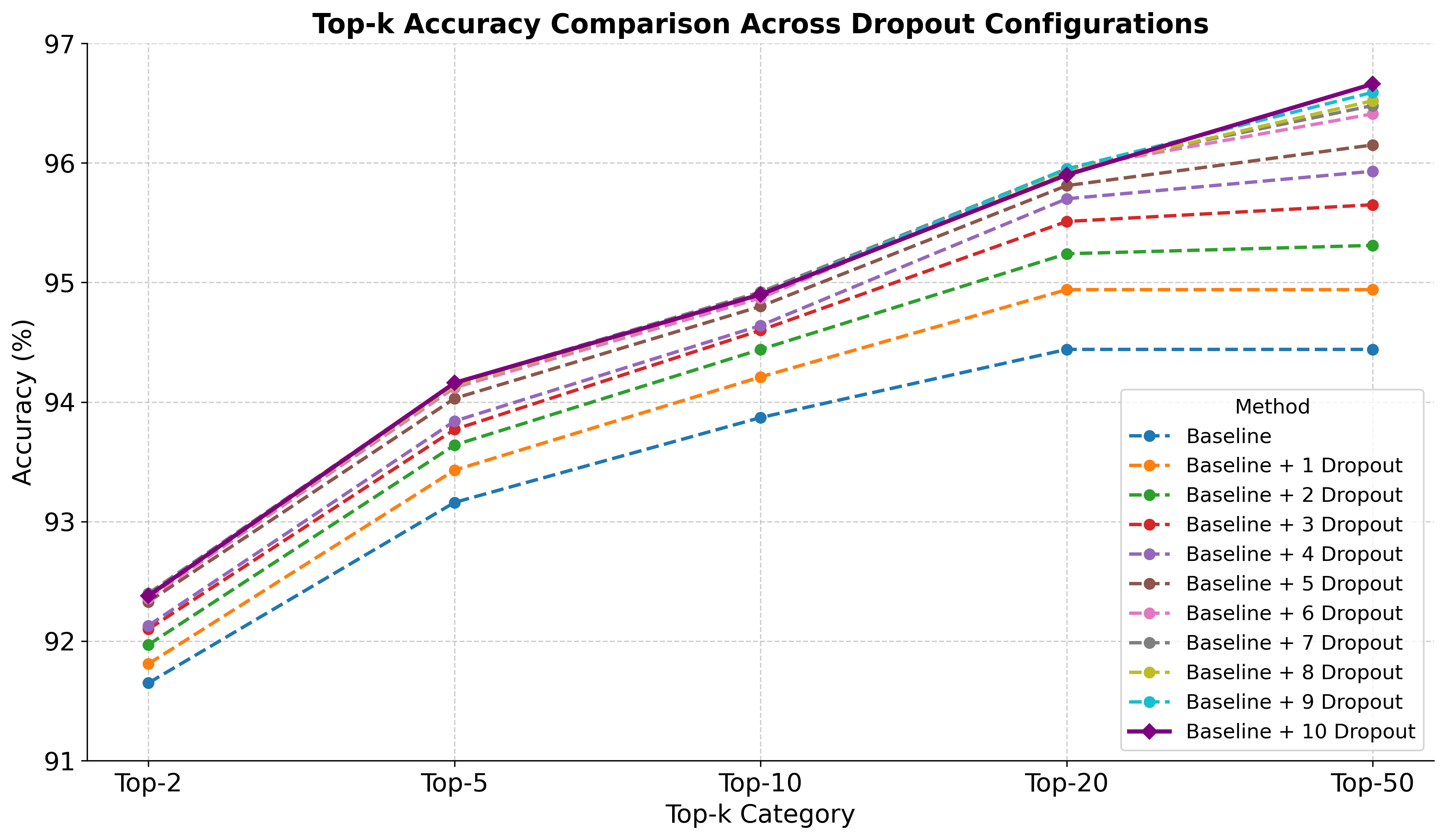}
\caption{Top-k accuracy improvements with an increasing number of Dropout runs. Accuracy improves steadily, with gains gradually diminishing beyond 7 dropouts.} 
\label{exp_dropout} 
\vspace{-0.15in}
\end{figure}

\paragraph{The impact of Inference-time dropout times}

Inference-time dropout is introduced to encourage divergent thinking.  We vary the number of dropout runs from 0 (baseline) to 10 (with different random seeds),  and report results in Fig.~\ref{exp_dropout}. 
A consistent performance improvement is observed as the number of dropout runs increases, with gains gradually saturating beyond 7 runs. This validates our designed inference-time dropout method can incentivize more small-scale divergent thinking of experts to generate diverse products.

\subsection{RQ3: Expert Behavior and Scoring}

\begin{figure*}
    \centering
    \begin{subfigure}{0.22\textwidth}
        \centering
        \includegraphics[width=\linewidth]{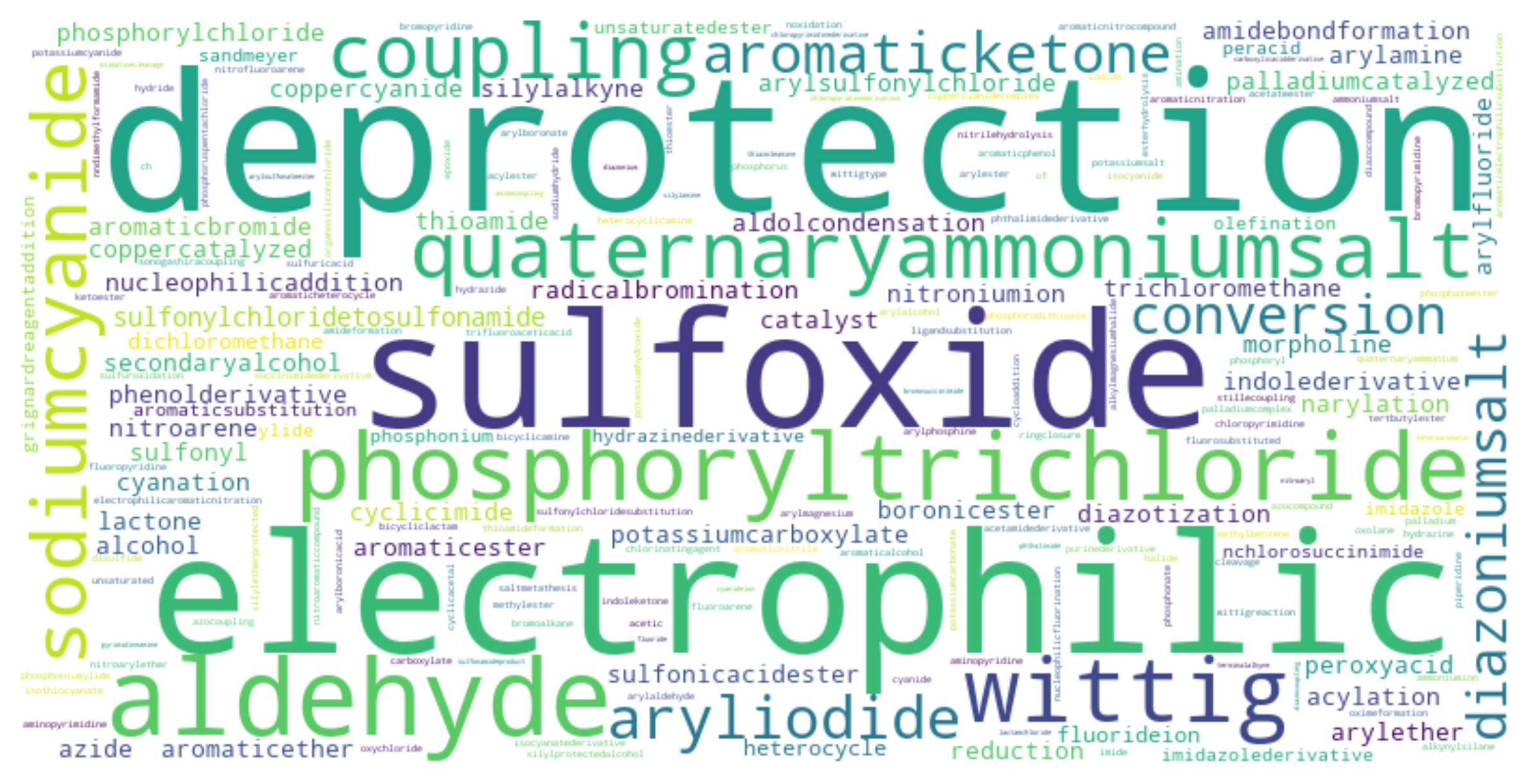}
        \caption{Chief Expert}
        \label{fig:sub1}
    \end{subfigure}
    \begin{subfigure}{0.22\textwidth}
        \centering
        \includegraphics[width=\linewidth]{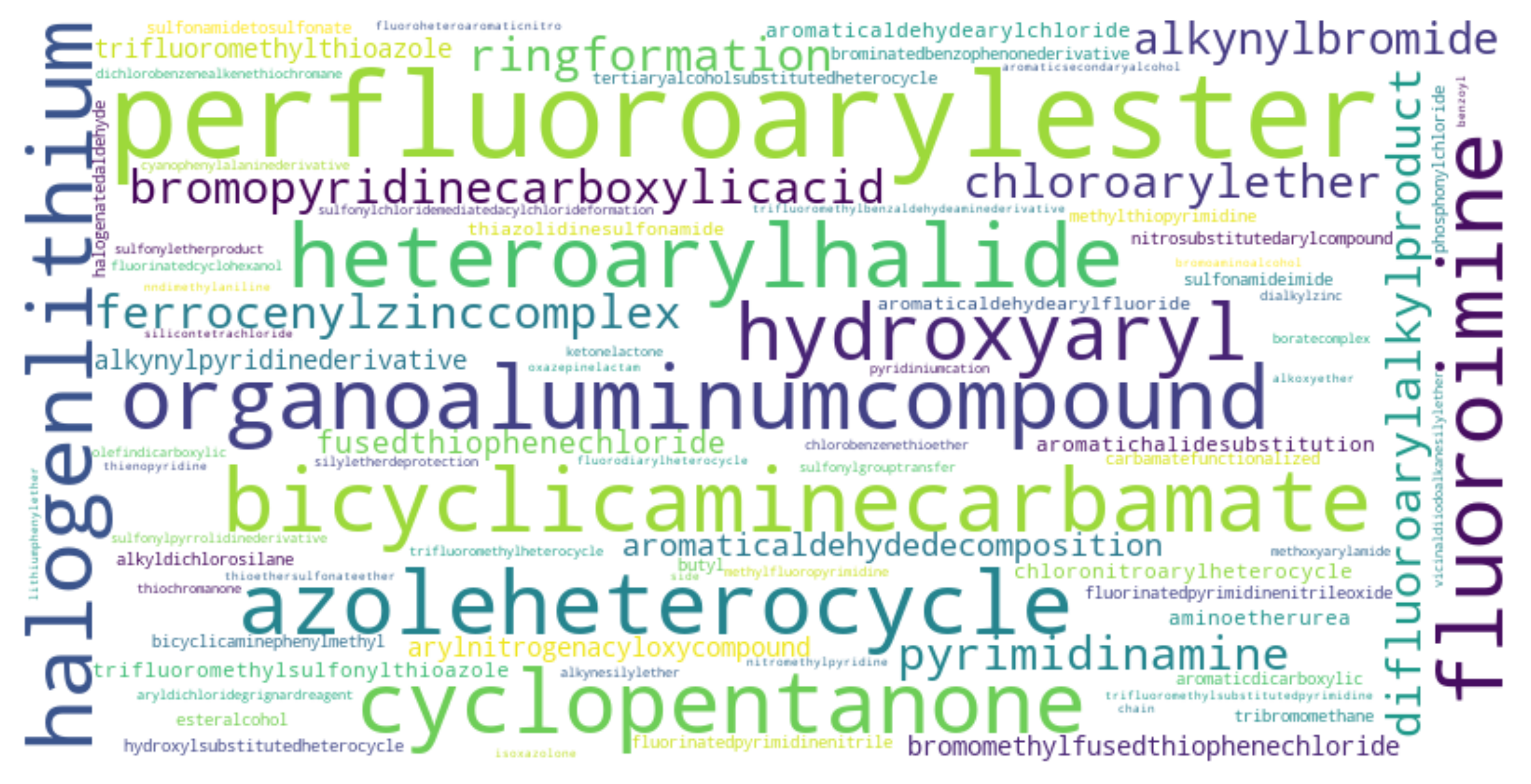}
        \caption{LoRA Experts}
        \label{fig:sub2}
    \end{subfigure}
    \begin{subfigure}{0.22\textwidth}
        \centering
        \includegraphics[width=\linewidth]{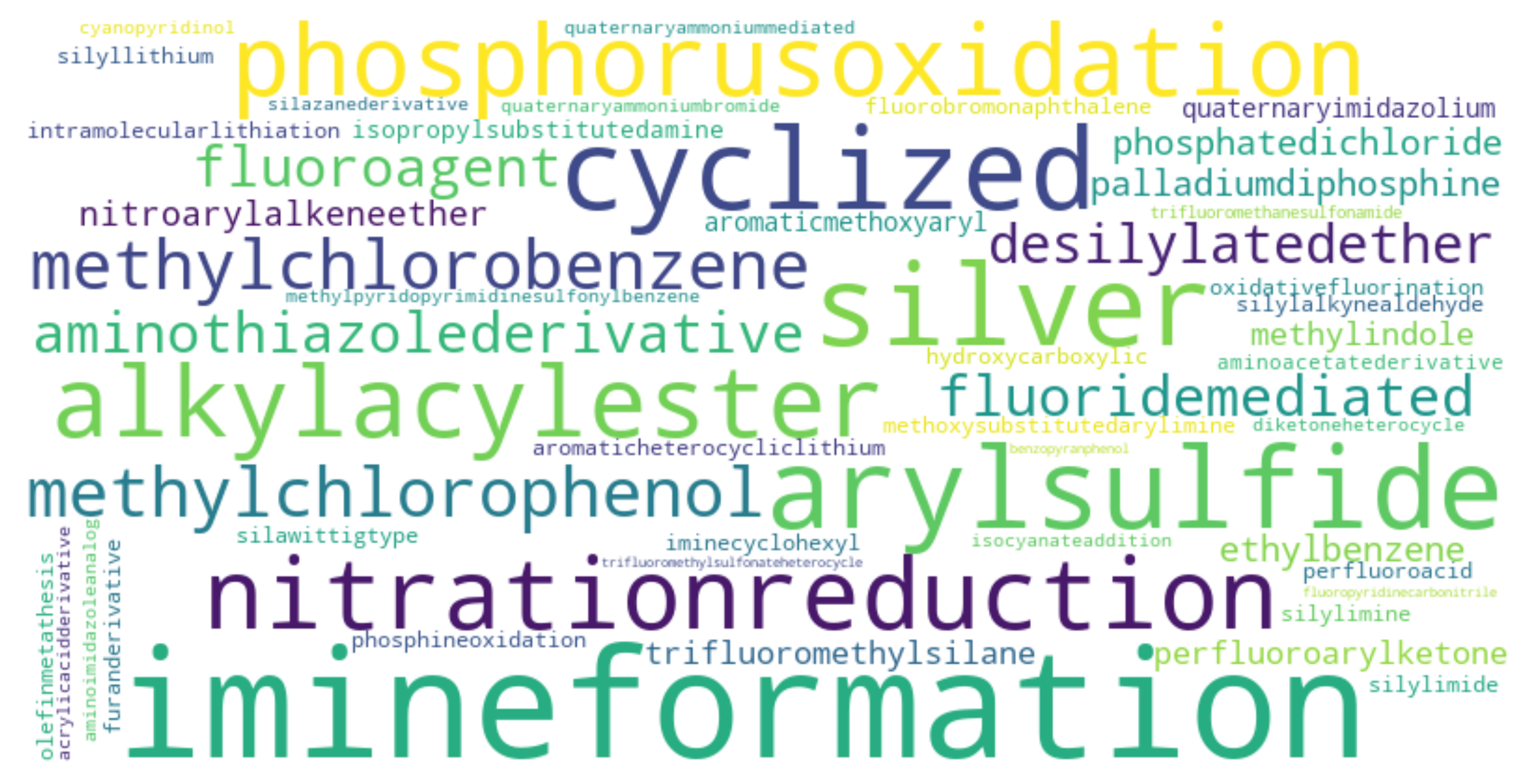}
        \caption{Chief Expert Dropout}
        \label{fig:sub3}
    \end{subfigure}
    \begin{subfigure}{0.22\textwidth}
        \centering
        \includegraphics[width=\linewidth]{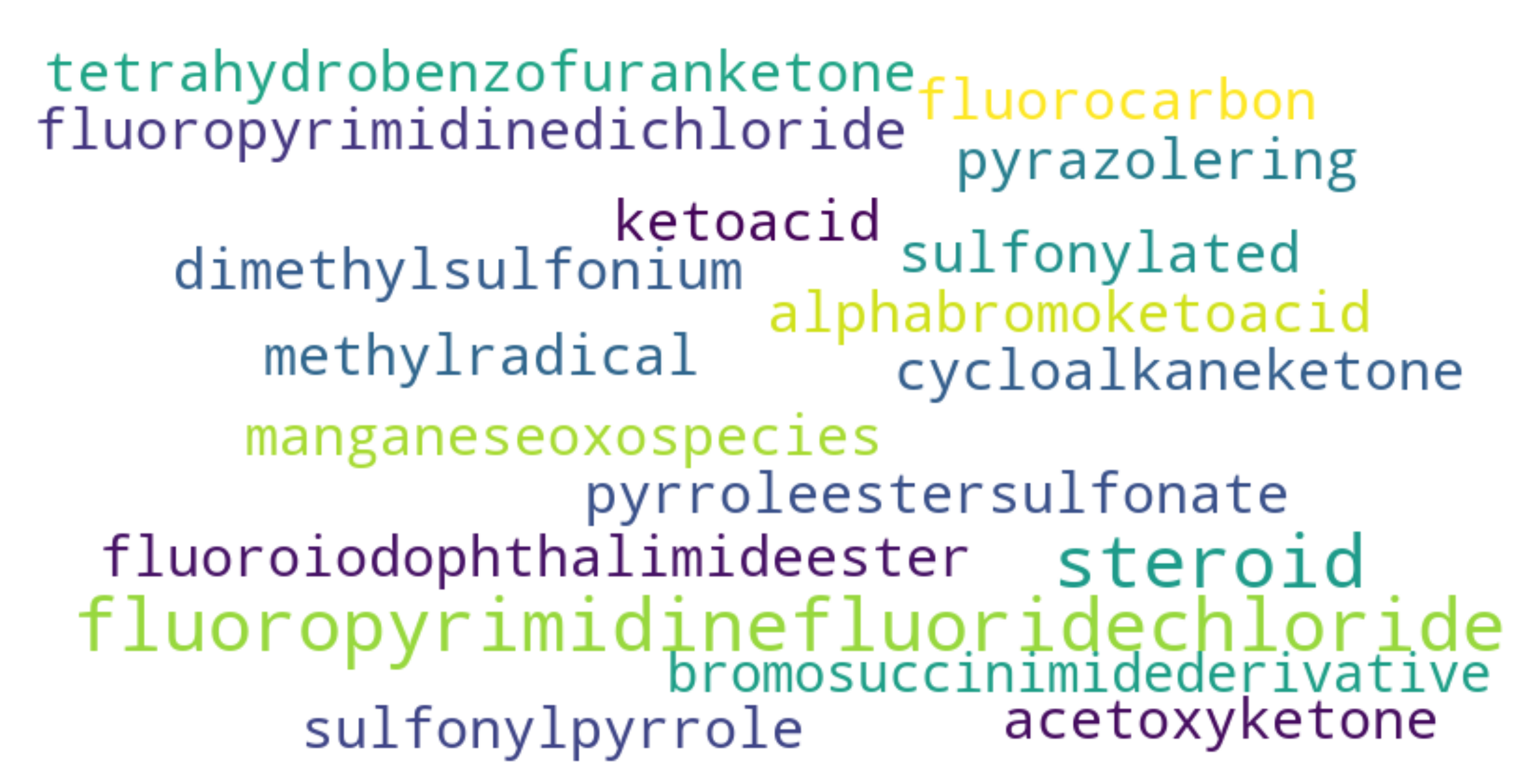}
        \caption{LoRA Experts Dropout}
        \label{fig:sub4}
    \end{subfigure}
    
    \caption{The reaction information explained by LLM based on the predictions of different experts. The Proportions of reaction types and related compounds differ across different experts and their variants.}
    \label{fig:exp_reac_types}
    \vspace{-0.05in}
\end{figure*}

\subsubsection{RQ3.1 How divergent experts handle distinct and atypical reaction patterns?}

While quantitative results show our approach excels, we next have a qualitative analysis to better understand how our divergent expert models behave in decoding for handling distinct reaction patterns.

\vspace{+0.1in}
\noindent \textbf{Handling Distinct Reaction Patterns:}
\begin{figure}
\centering
\vspace{-0.1in}
\includegraphics[width= 0.28\textwidth]{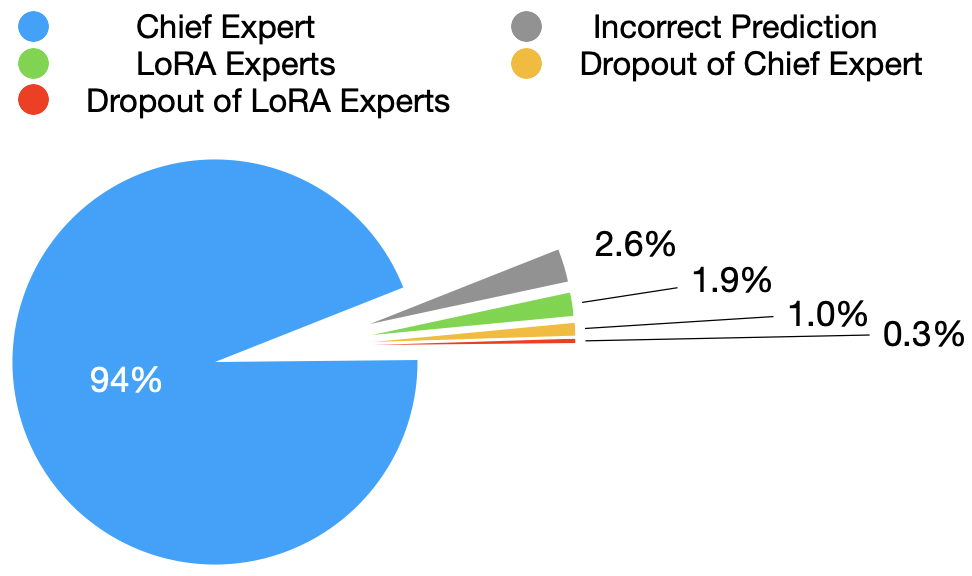}
\caption{Distribution of correct generations contributed by each expert (evaluated by Top-50 Accuracy), and the incorrect ones among the test set.} 
\label{fig:dis_exp_LoRAexperts} 
\vspace{-0.05in}
\end{figure}
To assess whether divergent experts specialize in different reaction patterns, we analyze correctly predicted samples among the test set. For each correct prediction, we identify which expert produced the output. \textcolor{black}{Fig. \ref{fig:dis_exp_LoRAexperts} distribution of correct generations contributed by each expert, as well as the incorrect ones. While 94\% of correct predictions are attributed to the chief expert, the LoRA experts and the dropout variants of both chief and LoRA experts also contribute to the overall performance.} We then input the corresponding reactants and predicted products into GPT-4.1~\cite{openai2023gpt4} to analyze the reaction type and the key compounds involved. This allows us to qualitatively assess the types of reactions each expert tends to specialize in. As shown in Fig.~\ref{fig:exp_reac_types}, GPT-generated descriptions reveal that these experts capture distinct reaction types and focus on different compound transformations. 
This confirms the effectiveness of our design in promoting divergent thinking across experts.

\begin{figure}
\centering
\includegraphics[width= 0.47\textwidth]{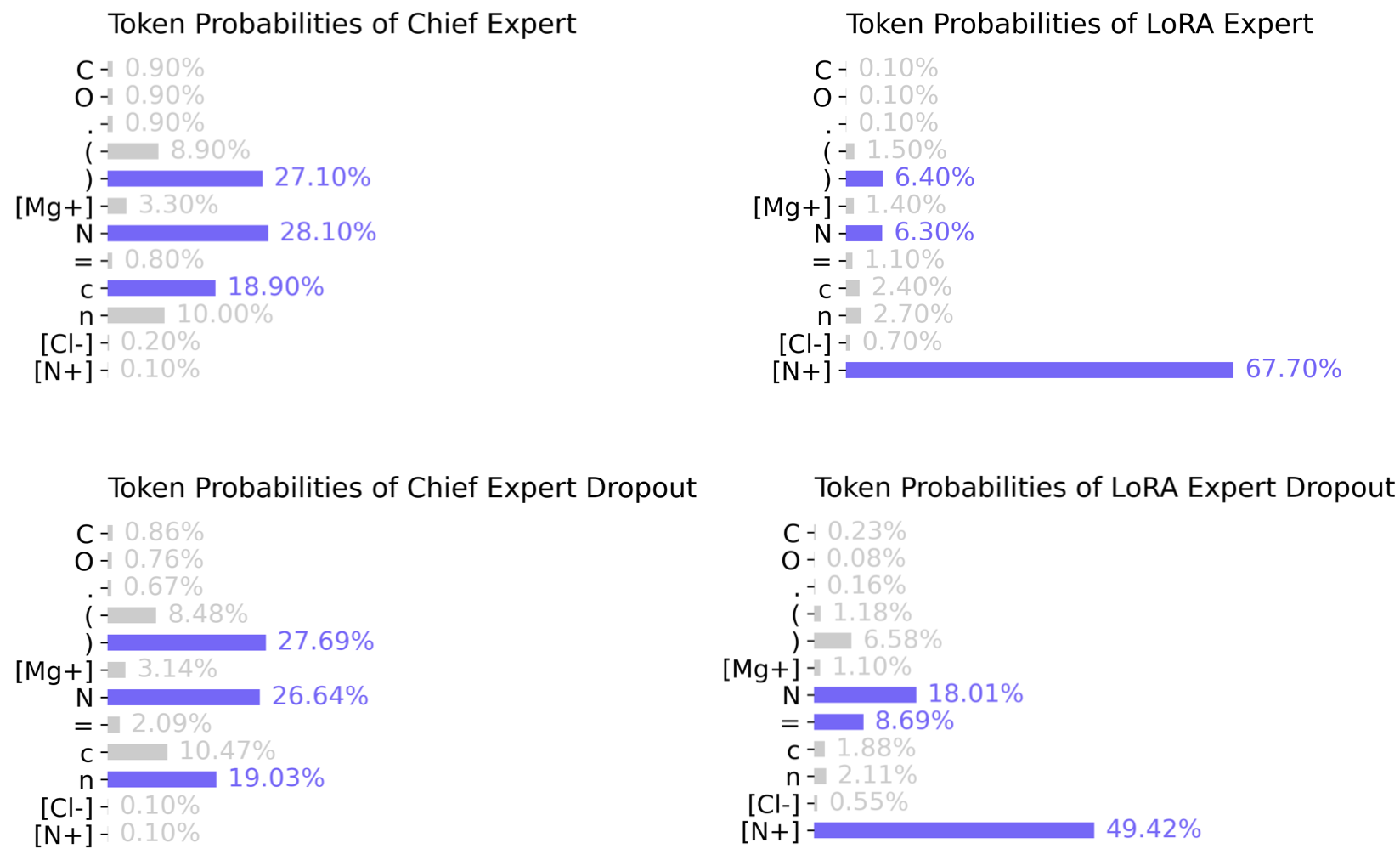}
\caption{Examples of token probabilities from different experts during decoding. The variation encourages diverse perspectives on reactions and leads to different product generations.} 
\label{fig:exp_token_prob} 
\vspace{-0.05in}
\end{figure}

\vspace{+0.1in}
\noindent \textbf{Influence of Divergent Thinking on the Decoding Process:}
To illustrate how divergent thinking emerges in the decoding process, we present a case study in Fig.~\ref{fig:exp_token_prob}. Given the same reactants, we compare the first-token probabilities from the chief expert, LoRA experts, and their dropout variants. The results show significant differences in token selection between the chief and LoRA experts, indicating \textbf{large-scale divergence} in their generative behavior. Moreover, we observe \textbf{small-scale divergence} between each expert and its dropout-based variants, suggesting that inference-time dropout encourages subtle yet meaningful variation in output. These findings reinforce our claim that both Seq. MOLE and inference-time Dropout mechanisms contribute complementary forms of divergence, which collectively enhance the model's capacity to generate plausible and diverse reaction outcomes.

\begin{figure*}
\centering
\includegraphics[width=0.64\textwidth]{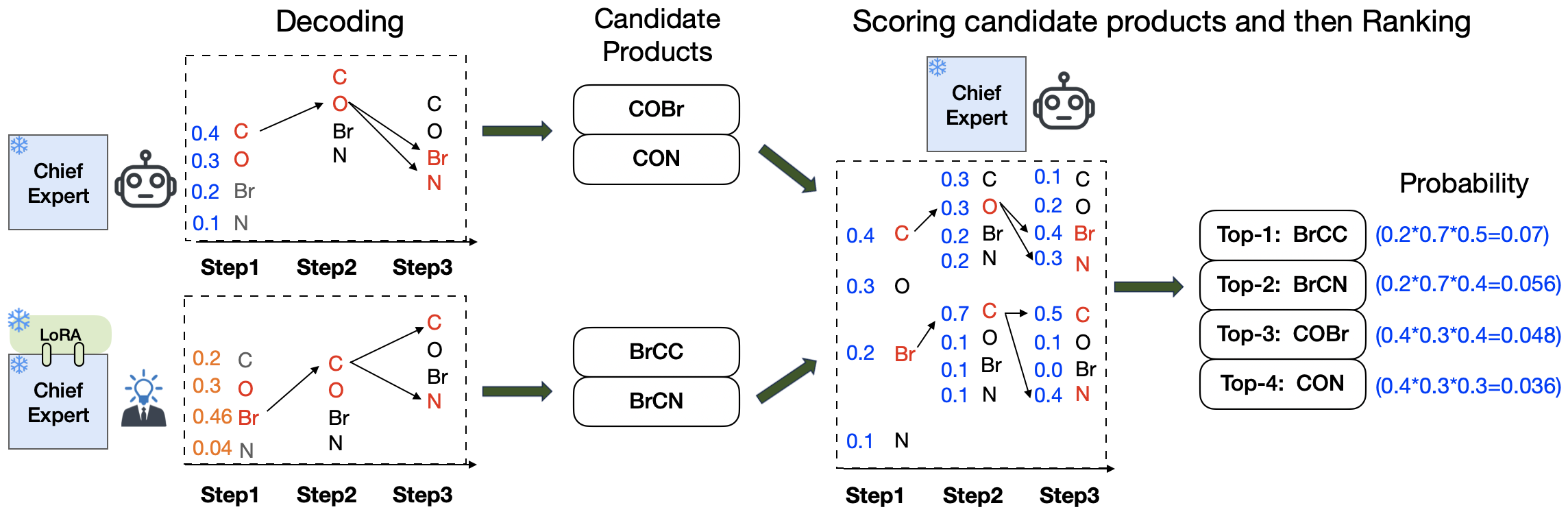}
\caption{One example illustrating why the chief model can be used for scoring and ranking (RQ2). Although the chief model ignores the \texttt{Br} token during beam search decoding with beam size 2,  it assigns a higher score to candidate products starting with \texttt{Br} generated by LoRA experts, when acting as an evaluator.} 
\label{exp_decoding} 
\vspace{-0.1in}
\end{figure*}

\subsubsection{RQ3.2 Why does it remain valid to use the $f_{\text{Chief}}$ for scoring and ranking?}
\label{sec:RQ2}

As discussed in Section \ref{sec:gating}, reusing  the chief expert for scoring and ranking is a practically reasonable solution. 
In this section, we investigate how the chief expert can enhance decoding performance.  One key insight drawn from our empirical evaluation is: \textbf{The chief expert is a qualified evaluator of candidate products, capable of assigning higher scores to outputs  generated by other specialized experts than to its own predictions}.

One case study is illustrated in Fig.~\ref{exp_decoding}, where we use a beam size of 2 and restrict each step to the top-2 next-token candidates. The chief expert may omit less likely initial tokens such as \texttt{Br}, \texttt{N} during its own generation process. In contrast, divergent LoRA experts, trained on different token distributions, may assign higher probabilities to these rare tokens, thus generating structurally different product candidates.
More importantly, although the chief expert assigns lower probability to these tokens (e.g., \texttt{Br}) in the first step, it may assign higher conditional probabilities to the subsequent tokens in those sequences. When computing the joint likelihood over the full product SMILES, the total score for such candidates generated by LoRA experts, can exceed that of the chief expert’s own outputs. This explains why our method not only improves Top-50 accuracy through greater diversity, but also significantly boosts Top-2 and Top-5 accuracy by recovering plausible products that traditional methods often overlook.

While one might attempt to capture such diversity by increasing the beam size or the number of considered tokens (Top-N) per step, this approach has a critical limitation in the chemical domain. Unlike natural language, where approximate correctness is often acceptable, chemical reaction prediction in SMILES format requires \textbf{Exact Match}, thus a single wrong token can invalidate the entire prediction. Consequently, simply increasing beam size or top-n often leads to a higher proportion of invalid outputs without substantially improving accuracy. This limitation is evident in Fig.~\ref{fig:exp_all}, where increasing the beam size to 50 under the chief expert yields only marginal improvement from Top-20 to Top-50 accuracy. This highlights that \textbf{traditional beam search fails to generate additional plausible} products despite increased diversity. In contrast, our set of divergent expert models acts as a form of structured subsearch: they explore alternative token trajectories that are chemically valid yet missed by the base model. When scored by the chief expert, these candidates represent a refined and more effective decoding strategy that emphasizes plausibility over pure diversity. Thus, 
\textbf{our divergent experts can be alternatively regarded as an effective subsearch mechanism for plausible products in the reaction space during decoding}.

\subsection{Broader Impacts}
\label{limit}
Our key contribution is being the first to highlight the importance of \textbf{beyond the typical} in reaction prediction. Prior methods emphasize only Top-1 performance, making models biased toward common patterns and suppresses novelty. Our work has the potential to benefit other science domains that seek \textbf{``atypical''} patterns, such as biology, material science, and so on. For instance, in biology, rare but functionally significant protein–ligand binding modes are often overlooked by models trained only on dominant interactions.

%% file: 5-Conclusion.tex
\section{Conclusion}
Reaction prediction is a fundamental challenge. We identify a major limitation ignored by existing methods: atypical patterns, and introduce \textit{divergent thinking} to address it. Our method assembles experts to generate diverse, plausible outcomes, mimicking reasoning of chemists. Experiments show it not only improves performance but also provides insights into the value of divergent thinking in reaction modeling.

\section{Acknowledgments and Disclosure of Funding}
This work was supported by 
the NSF Center for Computer-Assisted Synthesis (C-CAS), No. CHE-2202693.

%% file: 6-Appendix.tex
\appendix
\section{Appendix}
In this appendix, we give a detailed gradient-based analysis to clarify why rare reaction patterns are hard to learn and how our method overcomes this limitation.

\subsection{Problem Setup: similar input reactants $x$, very different products $y$}

Consider a chemical reaction region of inputs $\mathcal{R}$ where all $x$ are very similar.
In this region, labels follow
\[
\Pr(Y = y_{\text{major}} \mid X = x) = 1 - \varepsilon, 
\quad
\Pr(Y = y_{\text{rare}} \mid X = x) = \varepsilon,
\]
for all $x \in \mathcal{R}$, with small $\varepsilon \ll 1$.
Thus, we have ``similar $x$, but sometimes a totally different $y$''. Different from most natural language processing tasks that only focus on $y_{major}$ and ignore $y_{rare}$, in chemistry reaction prediction or some natural science areas, the $y_{rare}$ is more important since it can boost scientific discover for breakthrough.

\subsection{Why Standard LM training struggles}
Standard LM training minimizes the negative log-likelihood $\mathcal{L}(\theta) 
= \mathbb{E}_{(X,Y)}\big[-\log p_\theta(Y \mid X)\big]$. Restricted to $\mathcal{R}$, this loss is
\begin{align*}
\mathcal{L}_{\mathcal{R}}(\theta)
&= (1-\varepsilon)\,\mathbb{E}_{X \in \mathcal{R}}\big[-\log p_\theta(y_{\text{major}} \mid X)\big] \\
&\quad + \varepsilon\,\mathbb{E}_{X \in \mathcal{R}}\big[-\log p_\theta(y_{\text{rare}} \mid X)\big].
\end{align*}

Since $\varepsilon \ll 1$, the gradient of $\mathcal{L}_{\mathcal{R}}$ is
\begin{align*}
\nabla_\theta \mathcal{L}_{\mathcal{R}}(\theta)
&= (1-\varepsilon)\,\mathbb{E}_{X \in \mathcal{R}}
    \big[-\nabla_\theta \log p_\theta(y_{\text{major}} \mid X)\big] \\
&\quad + \varepsilon\,\mathbb{E}_{X \in \mathcal{R}}
    \big[-\nabla_\theta \log p_\theta(y_{\text{rare}} \mid X)\big] \\
\end{align*}

Since $\varepsilon \ll 1$, the gradient of $\mathcal{L}_{\mathcal{R}}$ is dominated by the majority term and the model learns one ``blurred'' distribution for all similar $x$, and the rare,
totally different label $y_{\text{rare}}$, is weakly represented.

\subsection{Our Sequential Training Effectively Scales up a Diverse Mixture of Experts}

Formally, our method removes the samples correctly solved by the first model $f_1$ and uses the remaining ones as the training set for the next model $f_2$. Let:
\[
c_{\text{major}} 
= \Pr\big(f_1 \text{ correct} \mid X\in\mathcal{R}, Y=y_{\text{major}}\big),
\]
\[
c_{\text{rare}} 
= \Pr\big(f_1 \text{ correct} \mid X\in\mathcal{R}, Y=y_{\text{rare}}\big).
\]

Therefore, the \emph{effective} probability of the rare label in the training
data for $f_2$ (restricted to $\mathcal{R}$) is
\[
\Pr_2(Y = y_{\text{rare}} \mid X \in \mathcal{R})
=
\frac{\varepsilon(1 - c_{\text{rare}})}
     {\varepsilon(1 - c_{\text{rare}}) + (1-\varepsilon)(1 - c_{\text{major}})}.
\]

Typically, $f_1$ fits the easy majority much better than the rare cases, so
$c_{\text{major}}$ is large and $1 - c_{\text{major}}$ is very small, while
$c_{\text{rare}}$ is small and $1 - c_{\text{rare}}$ is large. In that case,
$\Pr_2(Y = y_{\text{rare}} \mid X \in \mathcal{R})
\gg \varepsilon$, 
Thus, the second model $M_2$ is trained almost entirely on the \emph{rare}
patterns in $\mathcal{R}$, i.e., on ``similar $x$, totally different $y$'' cases
that $f_1$ failed to learn, and the gradient for the second model will push the second model to satisfy:

{\footnotesize
\[
\nabla_{\theta_2} \mathcal{L}_{2,\mathcal{R}}(\theta_2)
\approx \Pr_2(Y = y_{\text{rare}} \mid X \in \mathcal{R})\,
\mathbb{E}_{X \in \mathcal{R}}
\big[-\nabla_{\theta_2}\log p_{\theta_2}(y_{\text{rare}} \mid X)\big],
\]
}

which put more strong supervised signal for the rare datasets on the second expert.

\subsection{Why other scaling methods such as mixture-of-experts fail}

Our method scales up experts through \emph{sequential} training, yielding diverse experts that can \textbf{handle similar inputs $x$ using different ways of thinking}. Gated Mixture-of-Experts (MoE) pursues a similar goal, but in this section we show why a standard gated MoE typically fails in this setting.
A gated MoE with $K$ experts has the form:

\[
p_{\text{MoE}}(Y \mid X)
= \sum_{k=1}^K g_k(X;\phi)\,p_k(Y \mid X;\theta_k),
\]

where $g_k(X;\phi)$ are gating weights (router parameters $\phi$).  Let
$Z \in \{1,\dots,K\}$
denote the (latent) expert index selected by the gate, with
\[
p(Z = k \mid X) = g_k(X;\phi).
\]

As training converges, this implies that for $X \in \mathcal{R}$, where inputs are nearly identical and only the labels $Y$ differ, we have

\[
p(Z = k \mid X = x, Y = y_{\text{major}})
= p(Z = k \mid X = x)
= g_k(x;\phi),
\]
\[
p(Z = k \mid X = x, Y = y_{\text{rare}})
= p(Z = k \mid X = x)
= g_k(x;\phi).
\]

Hence, considering $x$ are nearly identical, we have:
\[
p(Z = k \mid X = x, Y = y_{\text{major}})
\;\approx\;
p(Z = k \mid X = x, Y = y_{\text{rare}}).
\]

In this local region the routing expert index $Z$ depends almost \textbf{only
on $X$, and barely on whether $Y$ is the rare or majority outcome}; the MoE cannot
use $Z$ to encode ``this is a rare mode'', so the rare pattern does not receive a
dedicated expert in $\mathcal{R}$.




